\documentclass[final,1p,times]{elsarticle}

\usepackage{amsmath,amssymb,amsthm}
\usepackage{graphicx}
\usepackage{booktabs}
\usepackage{multirow}
\usepackage[table]{xcolor}
\usepackage{makecell}
\usepackage{gensymb}
\usepackage{float}
\usepackage{multicol}
\usepackage{comment}
\usepackage{pdfpages}
\usepackage{svg}
\usepackage{pgfplots}
\usepackage{url}
\pgfplotsset{compat=1.18}
\usepackage{hyperref}
\usepackage[caption=false]{subfig}
\usepackage{amssymb}   
\usepackage{booktabs}  
\newcommand{\xmark}{\text{\sffamily X}} 

\usepackage{hyperref}

\journal{Engineering Applications of Artificial Intelligence}

\begin{document}

\begin{frontmatter}

\title{ \huge A Diffusion-Contrastive Graph Neural Network with Virtual Nodes for Wind Nowcasting in Unobserved Regions}


\author{Jie Shi} 
\ead{j.shi1@uu.nl}

\author{Siamak Mehrkanoon \corref{cor1}}
\ead{s.mehrkanoon@uu.nl}

\cortext[cor1]{Corresponding author}

\address{Department of Information and Computing Sciences, Utrecht University, Utrecht, The Netherlands}


\begin{abstract}
Accurate weather nowcasting remains one of the central challenges in atmospheric science, with critical implications for climate resilience, energy security, and disaster preparedness. Since it is not feasible to deploy observation stations everywhere, some regions lack dense observational networks, resulting in unreliable short-term wind predictions across those unobserved areas. Here we present a deep graph self-supervised framework that extends nowcasting capability into such unobserved regions without requiring new sensors. Our approach introduces “virtual nodes” into a diffusion and contrastive-based graph neural network, enabling the model to learn wind condition (i.e., speed, direction and gusts) in places with no direct measurements. Using high-temporal resolution weather station data across the Netherlands, we demonstrate that this approach reduces nowcast mean absolute error (MAE) of wind speed, gusts, and direction in unobserved regions by more than 30\% - 46\% compared with interpolation and regression methods. 
By enabling localized nowcasts where no measurements exist, this method opens new pathways for renewable energy integration, agricultural planning, and early-warning systems in data-sparse regions.
\end{abstract}

\end{frontmatter}

\section{Introduction}

Weather and climate nowcasting underpin modern societies. From anticipating storm surges and flight safety to managing renewable energy grids, accurate short-term predictions of wind are essential for both human safety and economic resilience \cite{bib1,bib2,bib3}. While advances in numerical weather prediction and machine learning have dramatically improved forecasts in data-rich regions, much of the world remains outside the reach of dense observational networks. Remote rural areas, developing regions, offshore zones, and mountainous terrain often lack the sensors necessary to capture local atmospheric dynamics. Most nowcasting models rely on dense sensor networks to monitor changes in wind patterns over time and space \cite{bib4}. As a result, the regions most exposed to climate risks are often those with the weakest access to timely and reliable nowcasts.

Efforts to overcome this limitation have typically relied on expanding sensor coverage. However, this approach is constrained by the difficulty and expense associated with building and maintaining weather stations. Automatic weather stations require consistent power, robust data connections, and frequent maintenance, which are typically unavailable in remote or low-income areas \cite{bib5}. In extreme locations such as offshore oceans, deserts, or mountains, the expense and logistical strain increase. As a result, many parts of the world still remain out the reach of extensive weather monitoring networks. This uneven distribution of ground-based sensors results in significant gaps in our nowcasting skills. An alternative approach is to develop models that use information from observed regions to generalize into unobserved regions by exploiting known geographical and atmospheric relationships. Such an approach would enable localized forecasts/nowcasts without requiring new measurement infrastructure.

Recent deep learning models have shown impressive predictive capabilities in weather nowcasting task in regions with abundant observational data \cite{price2025probabilistic,ham2023anthropogenic,trebing2021smaat,reulen2024ga,vatamany2025graph, abdellaoui2021symbolic, aykas2021multistream}. However, deep learning model's capacity to generalize to data-scarce or even no data regions is limited. Many deep learning models operate under the assumption of having access to rich spatiotemporal input data \cite{allen2025end,alzubaidi2023survey}. Their performance usually drops significantly when applied to areas lacking observations. This limitation emerges when models are trained on data from heavily instrumented locations, which reduces their ability to adapt to spatial heterogeneity and data scarcity in unmonitored regions.

To address these challenges, we propose a virtual contrastive deep graph self-supervised framework (ContraVirt) that extends nowcasting capability into such unobserved regions without requiring new sensors. ContraVirt extends wind nowcasting into regions without direct measurements by integrating real observations with “virtual” locations. By embedding these virtual nodes within a graph neural network using graph diffusion based approach \cite{gasteiger2019diffusion} and training them using contrastive learning strategies \cite{chen2020simple}, the system learns to infer wind dynamics in unobserved areas while maintaining physical consistency with observed regions. 

To further improve the model's representation of unobserved regions, we implement two contrastive learning \cite{chen2020simple} strategies: The multi-step contrastive learning, which encourages virtual nodes to align with those of nearest real stations at future time steps, explicitly capturing realistic short-term dynamics dictated by geographic proximity and atmospheric continuity. The augmented contrastive learning, which enhances the robustness of representations by contrast each node and its masked version as positive pairs. These self-supervised strategies enable the model to learn stable embeddings even in the absence of observational data.

The ContraVirt framework is grounded in fundamental geographic and atmospheric principles. Tobler’s First Law of Geography \cite{bib6} states that nearby regions tend to share more attributes than distant ones, reflecting spatial continuity. Neighboring areas also exhibit geographical homogeneity, often characterized by similar terrain, vegetation, and land use, which contribute to comparable weather conditions. In addition, local atmospheric circulation systems, such as coastal sea–land breezes and mountain–valley winds, further reinforce correlations in weather across neighboring locations. Building on these concepts, ContraVirt employs a graph-based architecture that introduces virtual nodes to represent regions lacking direct observations. These nodes learn from their neighboring real nodes, enabling them to develop robust self-representations and effectively \lq fill in the blanks\rq in the observational network.

ContraVirt provides a low-cost alternative to dense sensor networks, extending wind nowcasting into regions without direct observations. We demonstrate its effectiveness in the Netherlands, a coastal setting where wind fields are shaped by North Sea storms and pronounced seasonal variability \cite{la2025meteorological}. It shows substantial improvements over established baselines. More broadly, ContraVirt provides a foundation for delivering reliable nowcast to data-sparse regions where they are most needed.

\section{Related Works}\label{sec:2}
In recent years, deep learning techniques have found applications in various weather forecasting tasks \cite{mehrkanoon2019deep, chen2020deep, ravuri2021skilful,  fernandez2021broad,gao2021deep,yang2022aa,fernandez2022deep,upadhyay2024theoretical, trebing2020wind, stanczyk2021deep}, yet their coverage remains geographically limited, underscoring the need for additional methods in unobserved areas. Conventional interpolation techniques, such as inverse distance weighting (IDW) and kriging, assume smooth spatial continuity and often fail to represent abrupt meteorological transitions caused by complex terrain or coastal effects \citep{cressie1990kriging, shepard1968idw}. Data assimilation within numerical weather prediction (NWP) systems can theoretically propagate information into unobserved areas, but the resulting forecasts are heavily dependent on model resolution and parameterization schemes, which are computationally expensive to refine \citep{wilks2011statistical}. Moreover, the spatial relationships used for interpolation or message passing are often predefined by distance or static correlations, ignoring dynamic atmospheric connectivity that evolves over time. Recent advances in data-driven modeling have explored deep learning and graph-based neural network architectures using various data sources and advanced modeling techniques \cite{wu2022promoting,yang2022spatio}. However, most existing approaches remain limited by their reliance on observed features, restricting their ability to generalize to locations without direct measurements. 

Graph diffusion provides a principled framework for modeling how information, energy, or atmospheric signals propagate across interconnected spatial domains. Consider directional flow and higher-order spatial dependencies, allowing long-range dependencies to emerge naturally through learned diffusion processes \citep{gasteiger2019diffusion}. In meteorological applications, diffusion graph convolution has been adopted in air quality prediction, wind power estimation, and rainfall nowcasting, effectively capturing how local disturbances evolve and spread through the atmospheric network \cite{andrae2024continuous,dong2025double}. Recent efforts have begun exploring adaptive diffusion mechanisms or multi-scale graph propagation to better capture evolving weather patterns \cite{han2024multi}. However, the application of graph diffusion to unobserved regions and temporally varying graphs remains largely unexplored.

Contrastive learning \cite{chen2020simple}, as an unsupervised representation learning paradigm, holds great potential for addressing weather forecasting challenges in unobserved regions where labeled data are scarce or unavailable. Contrastive frameworks encourage models to discriminate similar and dissimilar structures without supervision \citep{kumar2022contrastive}. This mechanism is closely aligned with Tobler’s First Law of Geography \cite{bib6}, which posits that neighboring regions tend to exhibit similar atmospheric conditions, while areas farther apart often differ. More recent studies have introduced masked or augmented contrastive objectives to recover latent dynamics from partially observed fields \cite{huang2025cross, gong2024spatio}. Nevertheless, existing approaches rarely consider graph-structured meteorological networks in unobserved regions.

\begin{figure*}[h]
    \centering
    \subfloat[33 automatic weather station (AWS) locations in the Netherlands. Both blue and green nodes indicate real stations; blue ones are used for training, whereas green stations are removed out entirely and replaced by virtual nodes during training, serving as test targets only.\label{fig1a}]{
        \includegraphics[width=0.3\textwidth]{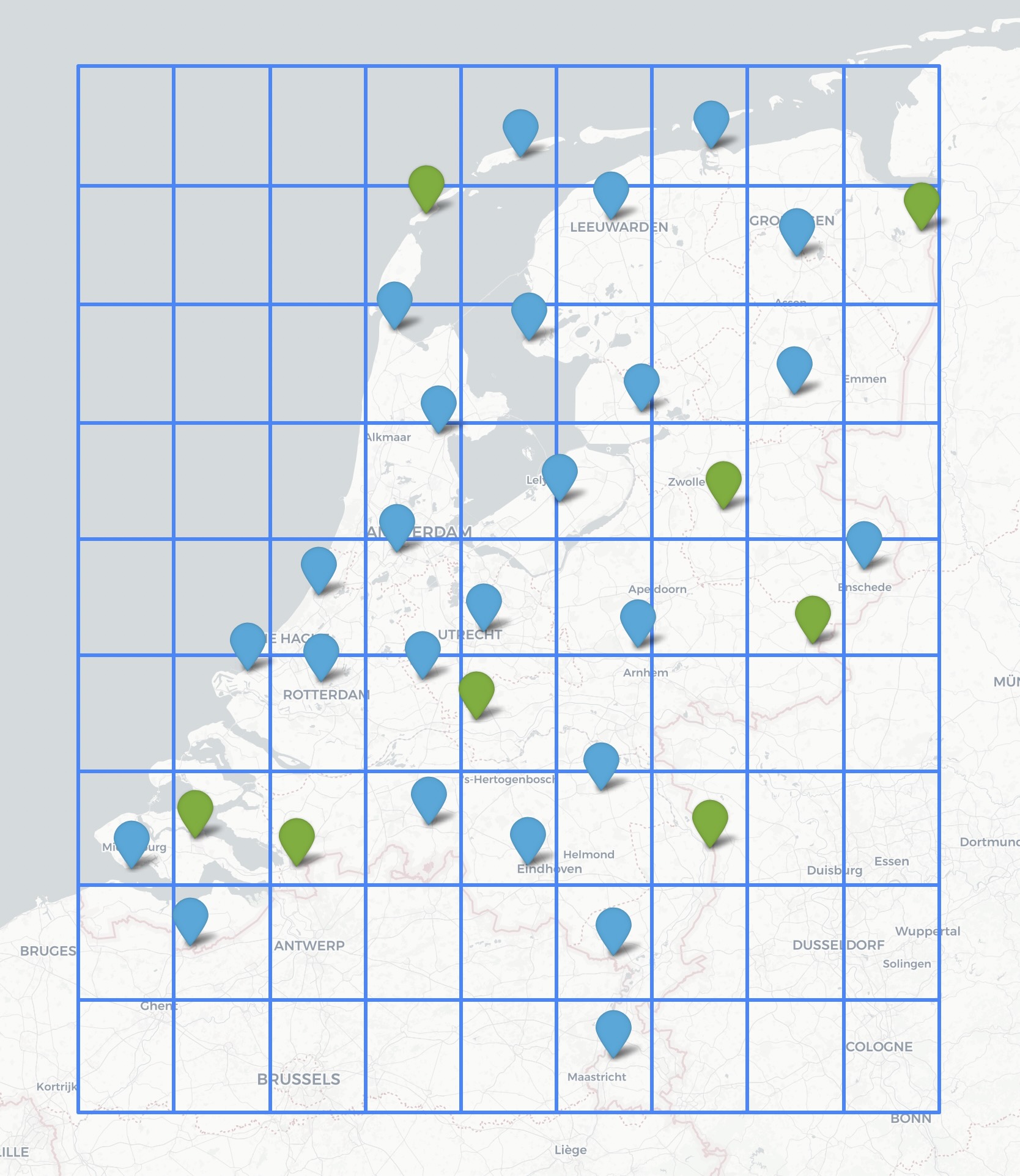}
    }\hfill
    \subfloat[Training nodes over a grid covering the entire Netherlands. Blue nodes represent real stations, limited to one per grid cell. Red nodes are virtual: either replacing excluded test stations or filling uncovered grid cells.\label{fig1b}]{
        \includegraphics[width=0.3\textwidth]{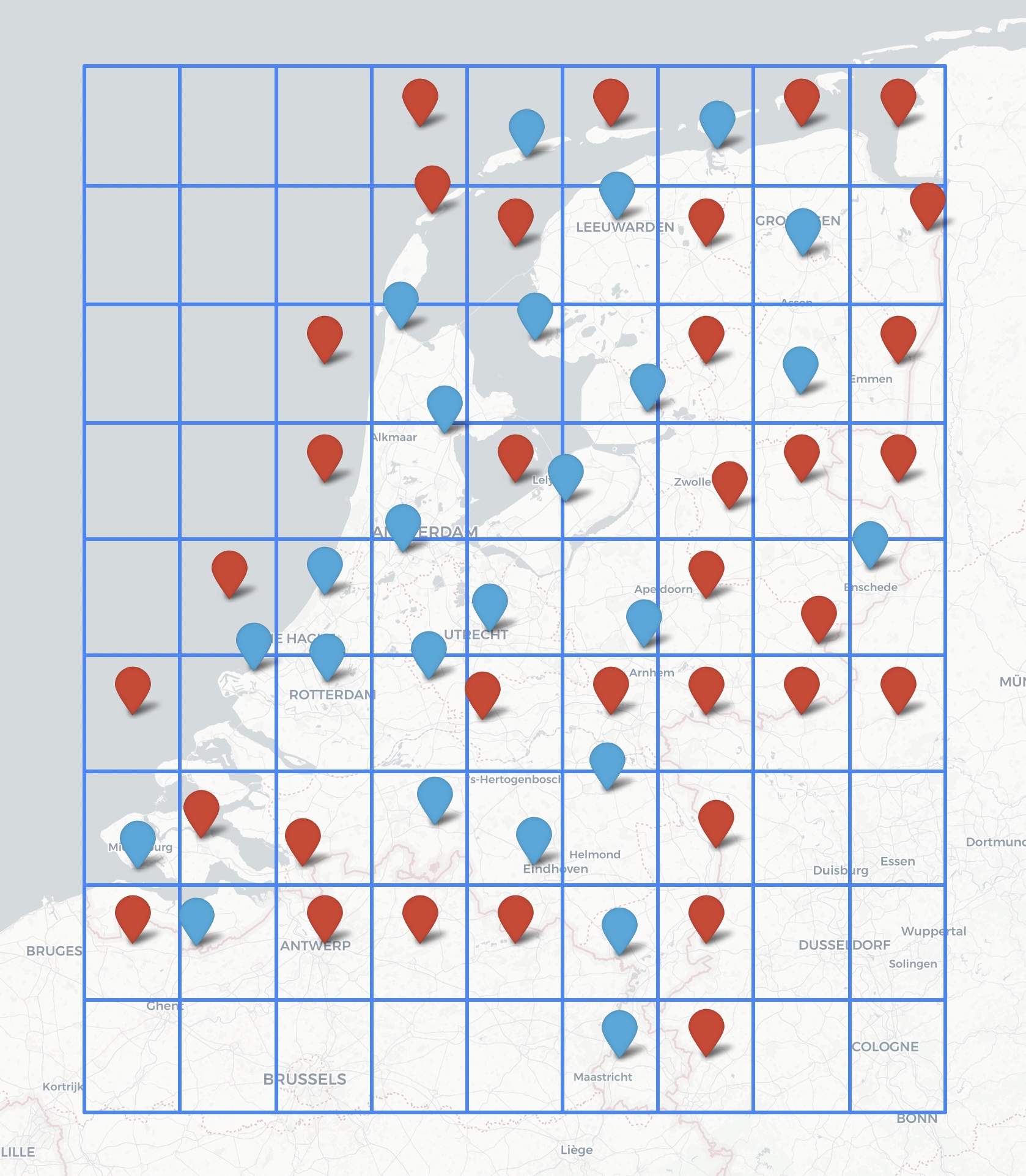}
    }\hfill
    \subfloat[Selected test stations, which are withheld from the training process. Model predictions generated via virtual nodes are evaluated using ground truth measurements from these real stations.\label{fig1c}]{
        \includegraphics[width=0.3\textwidth]{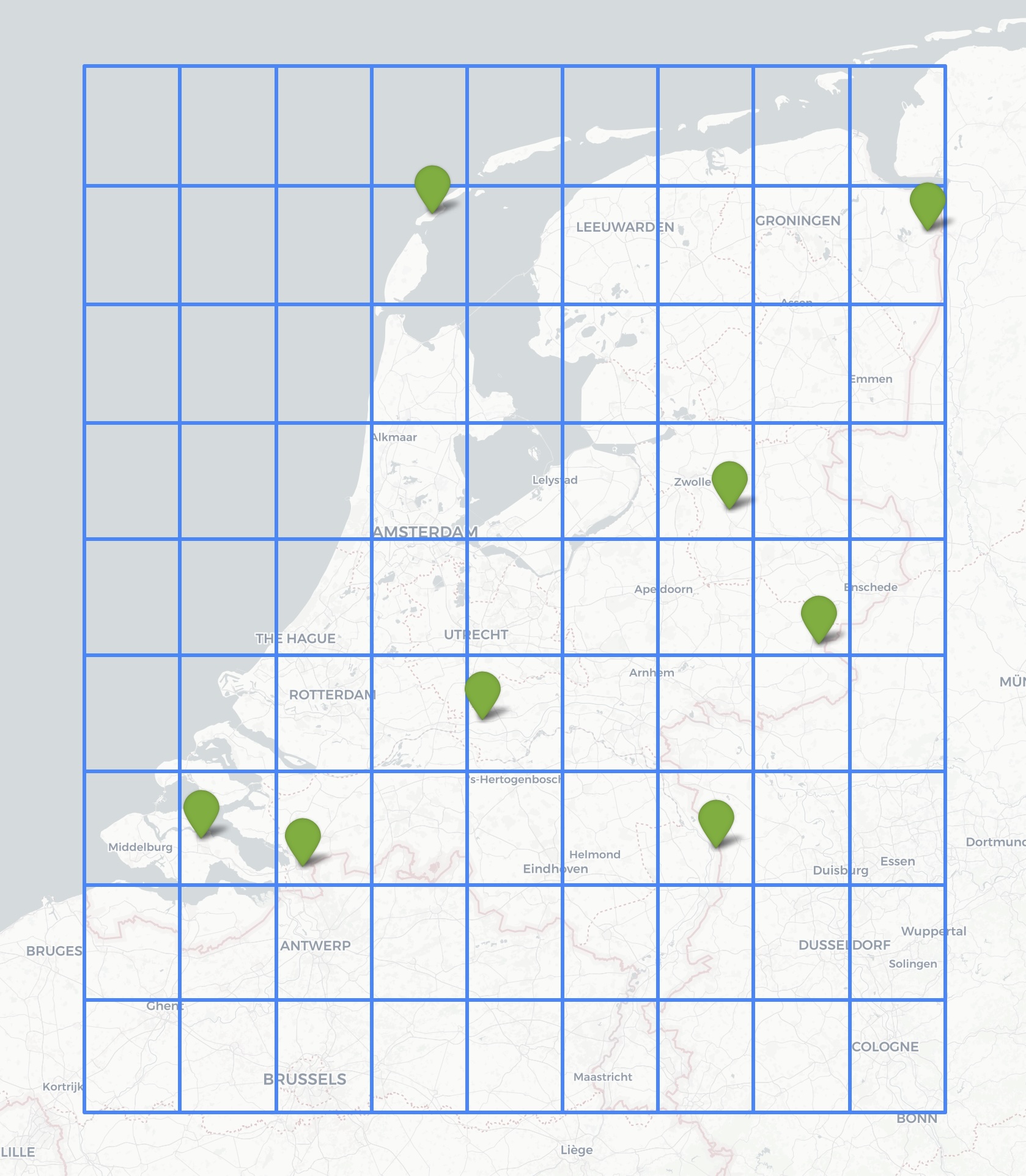}
    }
    \caption{Overview of meteorological station locations and the spatiotemporal coverage of training and testing datasets across the Netherlands.}
    \label{fig:1}
\end{figure*}
\section{Method}\label{sec:3}

\subsection{Virtual nodes}
ContraVirt model is a spatio-temporal graph neural networks that combines real meteorological stations with strategically placed virtual nodes. These virtual nodes serve as placeholders for unobserved regions. To ensure full coverage across the map, we begin by dividing the geographic domain into a fixed grid (Fig.\ref{fig1a}). Each grid cell is allowed to contain at most one real weather station. If no station is present, a virtual node is placed at the center of the cell. This results in a hybrid network of real and virtual locations. The real stations provide ground-truth observations, while virtual nodes enable the model to learn representations for regions without direct measurements.

In total, there are 34 automatic weather stations (AWS) located onshore across the Netherlands. However, the WIJK AAN ZEE AWS station (latitude: 52.5053, longitude: 4.6029) does not record wind-related variables. As a result, only 33 AWS stations were included in this study for wind nowcasting tasks. To ensure full spatial coverage across the Netherlands, a $9 \times 9$ grid map is constructed, with the constraint that each grid cell contains at most one real station. All 33 real AWS stations are located in distinct grid cells.

For grid cells without any real station, a virtual node is placed at the geometric center of the cell. These virtual nodes serve to fill spatial gaps and support the model’s generalization to unmonitored regions (see Fig.~\ref{fig1b}). Since virtual nodes lack both input features and ground truth observations, it is challenging to directly evaluate model performance at these locations. To rigorously evaluate the model’s performance on virtual nodes, a subset of real stations is intentionally excluded from training and replaced by virtual nodes with no input features. In the training phase, real weather stations are marked in blue, while virtual nodes are shown in red (see Fig.~\ref{fig1b}).  The virtual nodes consist of two types: (1) nodes that replace test stations, using the exact coordinates of the original test stations, and (2) nodes placed in grid cells without any station coverage, positioned at the geometric center of the corresponding cell. During inference, the model's predictions at these virtual node locations are compared against the actual ground truth from the corresponding real stations, allowing us to assess the model’s ability to generalize to data-sparse regions. 

As shown in Fig.~\ref{fig1c}, the 33 real stations are split into 25 training nodes and 8 testing nodes, represented by blue and green markers, respectively. Blue stations are included in the training set, while green stations are excluded from training and used exclusively for testing. During model training, each green station is replaced by a corresponding virtual node to simulate unseen locations.

As shown in Table \ref{tab:1}, of the 29 meteorological variables considered, wind direction (dd), wind speed (ff), and wind gust (gff) serve both as forecast targets and as lagged input features. Unlike real nodes, virtual nodes lack observations and thus have no direct input features. To approximate their meteorological variables, we compute weighted averages from the three nearest real stations. For lag features, which capture short-term atmospheric dynamics, we employ learnable embeddings \cite{bai2020adaptive} that enable the model to internally represent and adapt these features during training. This design allows the model to reconstruct wind patterns from surrounding context rather than being constrained by missing data. Each node is further augmented with geo-embeddings (latitude and longitude), temporal embeddings, and a node-type embedding, yielding a comprehensive representation for the graph.

\begin{table}
\centering
\footnotesize
\caption{List of meteorological variables used in this study}
\label{tab:1}
\renewcommand{\arraystretch}{1.05}
\begin{tabular}{p{0.8cm} p{5.2cm} p{0.8cm}}
\toprule
\textbf{Variable} & \textbf{Variable Name} & \textbf{Units} \\
\midrule
dd    & Wind direction (10-min mean) & degree \\
ff    & Wind speed at 10\,m (10-min mean) & m\,s\textsuperscript{-1} \\
gff   & Wind gust at 10\,m (10-min max) & m\,s\textsuperscript{-1} \\
ta    & Air temperature (1-min mean) & °C \\
rh    & Relative humidity (1-min mean) & \% \\
pp    & Air pressure at sea level (1-min mean) & hPa \\
zm    & Meteorological optical range (10-min mean) & m \\
qg    & Global solar radiation (10-min mean) & W\,m\textsuperscript{-2} \\
D1H   & Rainfall duration in last hour & min \\
dr    & Precipitation duration (10-min sum) & s \\
R6H   & Rainfall in last 6 hours & mm \\
R12H  & Rainfall in last 12 hours & mm \\
R24H  & Rainfall in last 24 hours & mm \\
rg    & Precipitation intensity (10-min mean) & mm\,h\textsuperscript{-1} \\
ss    & Sunshine duration & min \\
td    & Dew point temperature (1-min mean) & °C \\
Tgn   & Grass temperature (10-min min) & °C \\
Tgn6  & Grass temperature min (6\,h) & °C \\
Tgn12 & Grass temperature min (12\,h) & °C \\
Tgn14 & Grass temperature min (14\,h) & °C \\
Tn    & Air temperature min (10-min) & °C \\
Tn6   & Air temperature min (6\,h) & °C \\
Tn12  & Air temperature min (12\,h) & °C \\
Tn14  & Air temperature min (14\,h) & °C \\
Tx    & Air temperature max (10-min) & °C \\
Tx6   & Air temperature max (6\,h) & °C \\
Tx12  & Air temperature max (12\,h) & °C \\
Tx24  & Air temperature max (24\,h) & °C \\
ww-10 & Weather code (previous 10-min interval) & code \\
\bottomrule
\end{tabular}

\end{table}

\begin{figure}
  \centering
  \includegraphics[width=\textwidth]{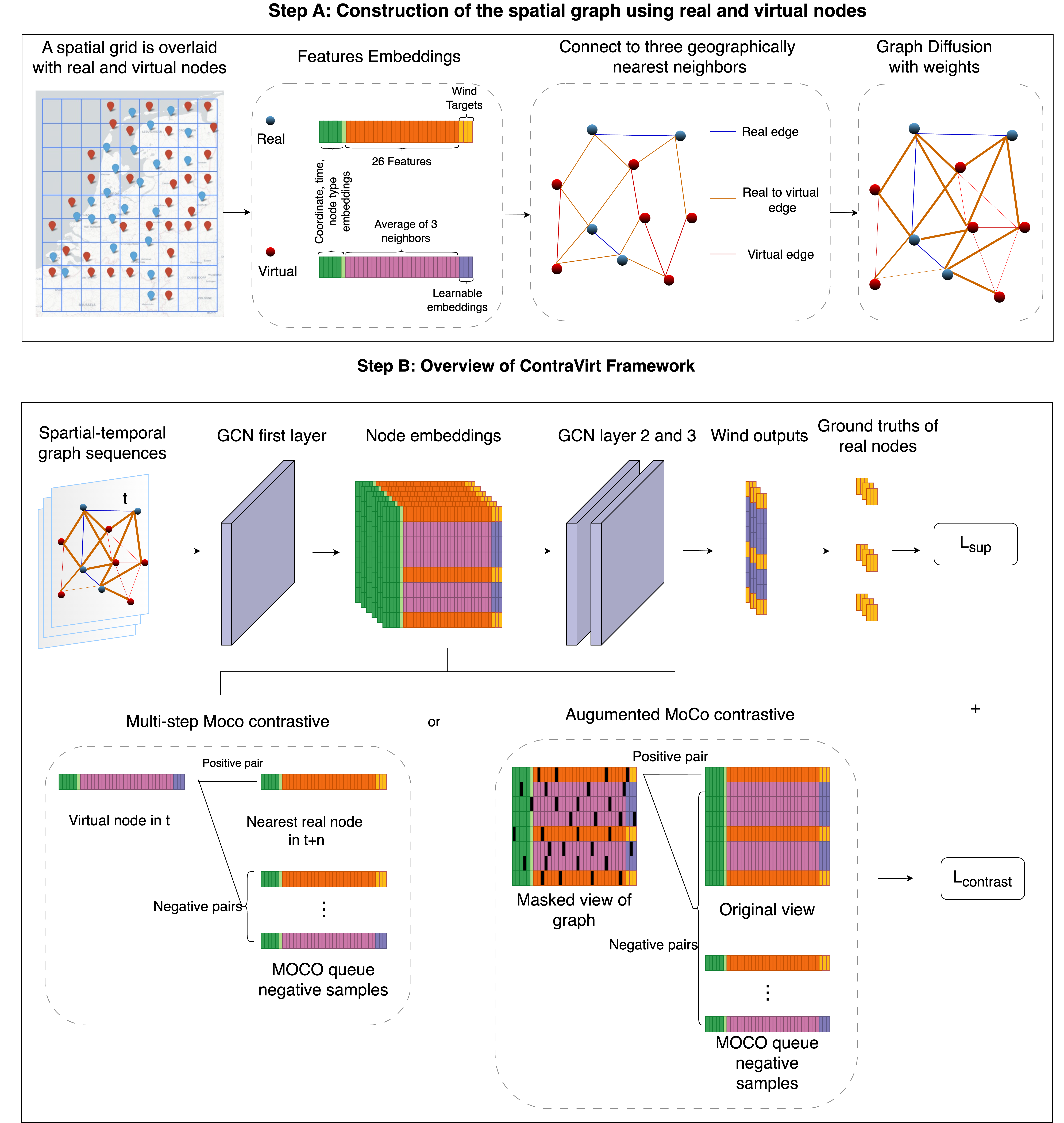}
  \caption{Overview of the model architecture and input graph generation process.
  Step A: Graph construction with real (blue) and virtual (red) nodes linked by diffusion-weighted edges. Step B: Model architecture combining supervised wind prediction on real nodes and self-supervised contrastive learning on virtual nodes.}
  \label{fig:2}
\end{figure}

\subsection{Model architecture}

We constructed the graph by linking each node to its three nearest neighbors based on geographic distance (Fig.\ref{fig:2}). The model follows the principles of Graph Neural Networks (GNNs), which exploit spatial proximity to enable neighborhood aggregation and improve node representations \cite{luan2023graph}. To extend beyond local interactions, we incorporate a graph diffusion mechanism based on Personalized PageRank \cite{page1999pagerank}, allowing weather dynamics to be propagated efficiently across the entire graph. We further introduce an edge-weighting strategy that prioritizes information transfer from real stations to virtual nodes while limiting exchange among virtual nodes. This design directs observational signals into unobserved regions, ensuring that virtual nodes receive sufficient information from nearby stations to support accurate nowcasting.

We frame the task as a temporal graph nowcasting problem targeting wind direction, speed, and gusts. The model (Fig.\ref{fig:2}) receives sequences of graphs sampled every 10 minutes over the previous 6 hours ($t-36$ to $t$) and predicts weather conditions up to 1 hour ahead ($t+1$ to $t+6$). This design aligns with nowcasting practice in operational nowcasting, focusing on the 0–1 hour horizon as the most critical and widely adopted timescale for high-resolution short-term forecasts. While prior studies often use multi-hour input windows, the exact number of steps varies across researches. Some short-term wind forecasting and nowcasting approaches rely on several hours of historical data to predict the next few hours (e.g., 4–6 hours ahead) \cite{zhang2025machine, chen2025ultra}. Hence, we adopt a 6 hours (36 steps) window that captures sufficient temporal dynamics without introducing outdated information.

Model training integrates two complementary objectives (Fig.\ref{fig:2}). For real stations, we use supervised regression to predict future wind conditions directly. For virtual nodes, which lack observations, we adopt a self-supervised contrastive learning \cite{chen2020simple} with two strategies. The first, a multi-step contrastive approach, aligns each virtual node at time $t$ with its nearest real station at a future time step. The second contrastive strategy is an augmented masking approach, which enhances representation learning by treating each virtual node and its augmented masked version as positive pairs. Specifically, we generate augmented versions of nodes by randomly masking input features, thereby encouraging the model to learn stable and generalizable embeddings even in the absence of explicit observational data.

To implement these contrastive strategies, we employ a momentum-based contrastive framework (MoCo) \cite{he2020momentum} that maintains a dynamic queue of negative samples, enhancing training stability and representation diversity. By combining supervised regression for real stations with contrastive self-supervision for virtual nodes, the model explicitly captures geographic and atmospheric dependencies, enabling accurate nowcasting in regions with sparse observations. This integrated design offers a scalable solution for wind nowcasting.

\subsubsection{Features embedding}
To model the spatiotemporal dependencies across both observed and unobserved regions, we construct a graph incorporating both real meteorological stations and virtual grid-based nodes. The feature processing begins with encoding each node’s spatial and temporal context. Specifically, for every node, we generate four geo-embeddings based on the geographical coordinates of each node, by applying sine and cosine transformations to both latitude and longitude values, a time-embedding capturing the periodicity of weather patterns, and a node-type embedding distinguishing real from virtual nodes. Real nodes are assigned 26 meteorological variables along with 3 wind-related lag features derived from historical observations. In contrast, virtual nodes lack direct observations. Their 26 variables are approximated using a weighted average of the three geographically nearest real nodes, while the lag features are represented by learnable embeddings to enable dynamic adaptation during training.

\subsubsection{Graph diffusion}
Next, a base graph is constructed by connecting each node to its three nearest spatial neighbors. We establish a base undirected graph $\mathcal{G}=(\mathcal{V}, \mathcal{E})$ where the node set $\mathcal{V}=\mathcal{V}_{r} \cup \mathcal{V}_{v}$, consists of both real nodes $\mathcal{V}_{r}$ and virtual nodes $\mathcal{V}_{v}$. Let $N=\left | \mathcal{V} \right | $ be the total number of nodes, and define the weighted adjacency matrix  $A \in \mathbb{R}^{N \times N}$. Let $\hat{A}=A+I$ and $\hat{D}$ be the corresponding degree matrix, where $I \in \mathbb{R}^{N \times N}$ denotes the identity matrix. The identity matrix plays a critical role in ensuring the invertibility of the expression and represents the probability of restarting the random walk from the source node in the diffusion process. We define the normalized transition matrix as follows, \cite{gasteiger2019diffusion} : 
\begin{equation}
    T=\hat{D}^{-1} \hat{A},
\end{equation}
On this initial structure, we apply Personalized PageRank (PPR)-based graph diffusion \cite{gasteiger2019diffusion} to enable global information propagation. The initial graph diffusion is computed via the Personalized PageRank (PPR) matrix \cite{gasteiger2019diffusion}:
\begin{equation}
\mathbf{D}_{\mathrm{PPR}}=\alpha_{p p r}\left(I-\left(1-\alpha_{p p r}\right) T\right)^{-1},
\end{equation}
where $\alpha_{p p r} \in(0,1)$ is the teleport probability \cite{chung2007heat}.
To further enhance information flow toward data-sparse regions, we introduce an edge reweighting strategy. We further reweight the diffusion matrix based on edge types. Specifically, for any pair of connected nodes $(i, j)$, the final diffusion weight $\tilde{d}_{i j}$ is computed as:
\begin{equation}
    \tilde{d}_{i j}=w_{i j} \cdot d_{i j}^{\mathrm{PPR}},
\end{equation}
where $d_{i j}^{\mathrm{PPR}}$ is the original $\mathbf{D}_{\mathrm{PPR}}$ value and $w{i j}$ is a type-specific scalar weight defined as:
\begin{equation}
    w_{i j}=\left\{\begin{array}{ll}
1, & \text { if } i, j \in V_{r} \\
\gamma, & \text { if } i \in V_{r}, j \in V_{v} \text { or } i \in V_{v}, j \in V_{r} \\
\delta, & \text { if } i, j \in V_{v},
\end{array}\right.
\end{equation}
This reweighted diffusion matrix allows for more targeted information propagation. It enhances the flow from real to virtual nodes $(\gamma>1)$, while damping interactions between virtual nodes $(\delta<1)$. This design encourages the model to prioritize the transmission of information from reliable sources (real nodes) to unobserved locations (virtual nodes), improving predictive performance in regions lacking measurements. In our implementation, we set $\gamma = 3$ and $\delta = 0.3$ to emphasize the importance of information flow from real observations to virtual locations while preventing excessive redundancy among virtual nodes. To ensure sparsity and computational efficiency, we retain only the top-8 entries per row in the reweighted diffusion matrix.

\subsubsection{Contrastive learning}
After constructing the diffusion-based graph, we define a spatiotemporal nowcasting task aimed at predicting wind direction, wind speed, and wind gust over the next six time steps, based on graph sequences from the preceding 36 time steps. Each graph in the sequence is first processed by a shared Graph Convolutional Network (GCN) \cite{zhang2019graph} layer to generate node embeddings. Let $h_i^t \in \mathbb{R}^d$ denote the embedding of node $i$ at time step $t$, obtained from the first GCN layer. These embeddings are then utilized for two parallel learning objectives: a self-supervised contrastive learning task involving both real and virtual nodes, and a supervised learning task applied to real nodes.

First, a self-supervised contrastive learning objective is applied to both real and virtual nodes, aiming to capture temporal consistency and spatial relationships. To explore different contrastive strategies, we adopt two complementary approaches. 
In the augmented contrastive strategy, we apply random feature masking to node representations and treat the masked and unmasked versions of the same node as a positive pair. All other nodes in the graph are considered negative samples. In the multi-step contrastive strategy, we treat each virtual node at time $t$
and its geographically  nearest real node at time $t+3$ as a positive pair, while all other combinations are treated as negatives. To further enhance training stability and diversity of negative samples, we adopt a momentum contrastive learning mechanism (MoCo) \cite{he2020momentum}, maintaining a dynamic memory queue of past embeddings to provide a rich and consistent set of negative samples across batches.

We define the similarity between two embeddings using cosine similarity:
\begin{equation}
\text{sim}(a, b) = \frac{a^\top b}{\|a\| \cdot \|b\|},
\end{equation}
where $\| \cdot \|$ denotes the $\ell_2$ norm.

In the Augmented Contrastive Strategy, for each node $i$, a masked view $\tilde{h}_i^t$ is generated by randomly masking part of its features. The contrastive loss \cite{chen2020simple} is defined as:
\begin{equation}
\mathcal{L}_{\text{aug}}^{(i)} = -\log \frac{\exp(\text{sim}(h_i^t, \tilde{h}_i^t)/\tau)}{\sum\limits_{j \in \mathcal{Q}} \exp(\text{sim}(h_i^t, h_j)/\tau)},
\end{equation}
The temperature parameter $\tau$ controls the concentration level of similarity scores within the softmax function. Smaller values of $\tau$ result in sharper distinctions between positive and negative pairs. where $\mathcal{Q}$ is a MoCo queue that stores negative samples.

In the Multi-step Contrastive Strategy, let $v_i^t \in V_v$ be a virtual node at time $t$, and let $r_i^{t+3} \in V_r$ denote its spatially nearest real node at time $t+3$. The contrastive loss \cite{chen2020simple} is defined as :
\begin{equation}
\mathcal{L}_{\text{multi}}^{(i)} = -\log \frac{\exp(\text{sim}(h_{v_i}^t, h_{r_i}^{t+3})/\tau)}{\sum\limits_{j \in \mathcal{Q}} \exp(\text{sim}(h_{v_i}^t, h_j)/\tau)},
\end{equation}

\noindent
Depending on the strategy, the total contrastive loss is computed as \cite{chen2020simple}:
\begin{equation}
\mathcal{L}_{\text{contrast}} = \frac{1}{N} \sum_{i=1}^{N} \mathcal{L}_{\text{aug}}^{(i)} \quad \text{or} \quad 
\mathcal{L}_{\text{contrast}} = \frac{1}{N'} \sum_{i=1}^{N'} \mathcal{L}_{\text{multi}}^{(i)},
\end{equation}
where $N$ denotes the total number of nodes in the batch for the augmented strategy, and $N'$ denotes the number of valid virtual nodes used in the multi-step strategy.

\subsubsection{Supervised objective}
A supervised nowcasting objective is applied exclusively to real nodes. Their embeddings are processed by two additional GCN layers and a regression head to predict future weather variables. The supervised component is trained using mean squared error (MSE) loss \cite{karunasingha2022root}. The total training objective combines the contrastive loss and the supervised loss, enabling the model to benefit from both labeled observations and unlabeled virtual information.

Let $\hat{y}_i^{t+1:t+6}$ denote the predicted future values of real node $i$, and $y_i^{t+1:t+6}$ the ground truth. The supervised loss is defined as \cite{karunasingha2022root}:
\begin{equation}
\mathcal{L}_{\text{sup}} = \frac{1}{|V_r|} \sum_{i \in V_r} \left\| \hat{y}_i^{t+1:t+6} - y_i^{t+1:t+6} \right\|_2^2.
\end{equation}

\subsubsection{Final training objective}
The total training loss is a combination of the supervised and contrastive objectives:
\begin{equation}
\mathcal{L}_{\text{total}} = \mathcal{L}_{\text{sup}} + \lambda \cdot \mathcal{L}_{\text{contrast}},
\end{equation}
where $\lambda$ is a hyperparameter that balances the contrastive objective against the supervised loss. To balance the influence of the contrastive loss during training, we define a dynamic weighting coefficient $\lambda$ based on a warm-up schedule and a sigmoid gating function. The final contrastive loss weight at each epoch is computed as:

\begin{equation}
\lambda = \lambda_{\text{warmup}} \cdot \lambda_{\text{mae}},
\end{equation}
We apply a linear warm-up over the first $E_{\text{warmup}}$ epochs:

\begin{equation}
\lambda_{\text{warmup}} = \min\left(1.0, \frac{e}{E_{\text{warmup}}}\right),
\end{equation}
where $e$ is the current epoch and $E_{\text{warmup}} = 20$. We further modulate the contrastive loss weight using a sigmoid function based on the previous epoch's MAE:

\begin{equation}
\lambda_{\text{mae}} =
\begin{cases}
0, & \text{if } e = 0 \\
\frac{1}{1 + \exp\left(-\kappa \cdot (\text{MAE}_{e-1} - \theta)\right)}, & \text{otherwise},
\end{cases}
\end{equation}
here, $\kappa = 10$ is a scaling factor controlling the sharpness of the sigmoid transition, and $\theta = 2$ is a predefined MAE threshold. This design enables contrastive learning to become increasingly influential only when the supervised signal stabilizes.

\section{Experiments}\label{sec:4}
\subsection{Dataset}
In total, there are 34 automatic weather stations (AWS) distributed across onshore areas of the Netherlands. However, the WIJK AAN ZEE AWS station (latitude: 52.5053, longitude: 4.6029) does not record wind-related variables. As a result, only 33 AWS stations were included in this study for wind nowcasting tasks. These processed data only include automated weather station (AWS) sites located onshore within the Netherlands. The original raw dataset provided by KNMI, which also contains wind poles and stations located on the BES islands, aerodromes, and North Sea platforms that were not utilized in this study, can be accessed at \url{https://dataplatform.knmi.nl/dataset/actuele10mindataknmistations-2}.

\subsection{Baselines}
The full model incorporates two contrastive strategies, multi-step and augmented, each strengthened by a MoCo-based queue that provides diverse negative samples, and is trained on a diffusion-based graph representation. To assess the contribution of each component, we performed an ablation analysis targeting the contrastive module, the MoCo enhancement, and the graph diffusion mechanism. First, we removed the MoCo queue while retaining both contrastive strategies. Next, we eliminated the contrastive module entirely, leaving the model without self-supervised constraints. Finally, we removed the diffusion module, reducing the architecture to a standard Graph Convolutional Network (GCN). These ablations isolate the role of each component and quantify their individual impacts on nowcasting accuracy.

In addition to the ablation analysis, we benchmarked our approach against four non-graph baselines: auto-regression (AR) \cite{ar}, linear regression (LR) \cite{lr}, K-nearest neighbors (KNN) \cite{knn}, and inverse distance weighting (IDW) \cite{idw}. AR and LR were implemented as multistep time-series predictors, taking the past 36 observations from the three nearest stations to forecast the next six steps at a given virtual node. In contrast, KNN and IDW are spatial interpolation methods that do not model temporal dependencies. For these, forecasts at each time step were estimated by aggregating simultaneous observations from the three closest stations: IDW applied inverse-distance weights, while KNN used a simple average. Extended Data Table 2 summarizes all tested models and their components, including detailed experimental settings.

\begin{table}
\centering
\setlength{\tabcolsep}{3pt}
\renewcommand{\arraystretch}{1.1}
\caption{Ablation and baseline model comparison. 
Each column denotes whether the corresponding component is included in the model.}
\label{tab:2}
\begin{tabular}{lcccc}
\toprule
\textbf{Model} & \textbf{Diffusion} & \textbf{Contrastive} & \textbf{MoCo} & \textbf{Self-supervised} \\
\midrule
ContraVirt (Augmented MoCo) & \checkmark & \checkmark & \checkmark   & \checkmark \\
ContraVirt (Multi-step MoCo) & \checkmark & \checkmark & \checkmark  & \checkmark \\
Augmented & \checkmark & \checkmark & \xmark  & \checkmark \\
Multi-step & \checkmark & \checkmark & \xmark  & \checkmark \\
\midrule
w/o Contrastive & \checkmark & \xmark & \xmark &\checkmark \\
w/o Contrastive \& Diffusion & \xmark & \xmark & \xmark  & \xmark \\
\midrule
AR & \xmark & \xmark & \xmark  & \xmark \\
LR & \xmark & \xmark & \xmark  & \xmark \\
KNN & \xmark & \xmark & \xmark & \xmark \\
IDW & \xmark & \xmark & \xmark  & \xmark \\
\bottomrule
\end{tabular}
\end{table}

\subsection{Training}
The models were implemented and trained using PyTorch. Training was conducted using the AdamW optimizer with a learning rate of 0.001. Early stopping was employed to mitigate overfitting and to enhance training stability. Additionally, a learning rate scheduler was used to reduce the learning rate when the training loss plateaued. Models were trained for up to 200 epochs with a batch size of 32. These hyperparameters were selected through preliminary experiments to balance nowcasting performance and computational efficiency. For augmentation-based contrastive learning, a masking ratio of 0.3 was used. The memory queue size for MoCo was set to 512.

\begin{figure}[!h]
    \centering
    \subfloat[]{%
        \includegraphics[width=\textwidth]{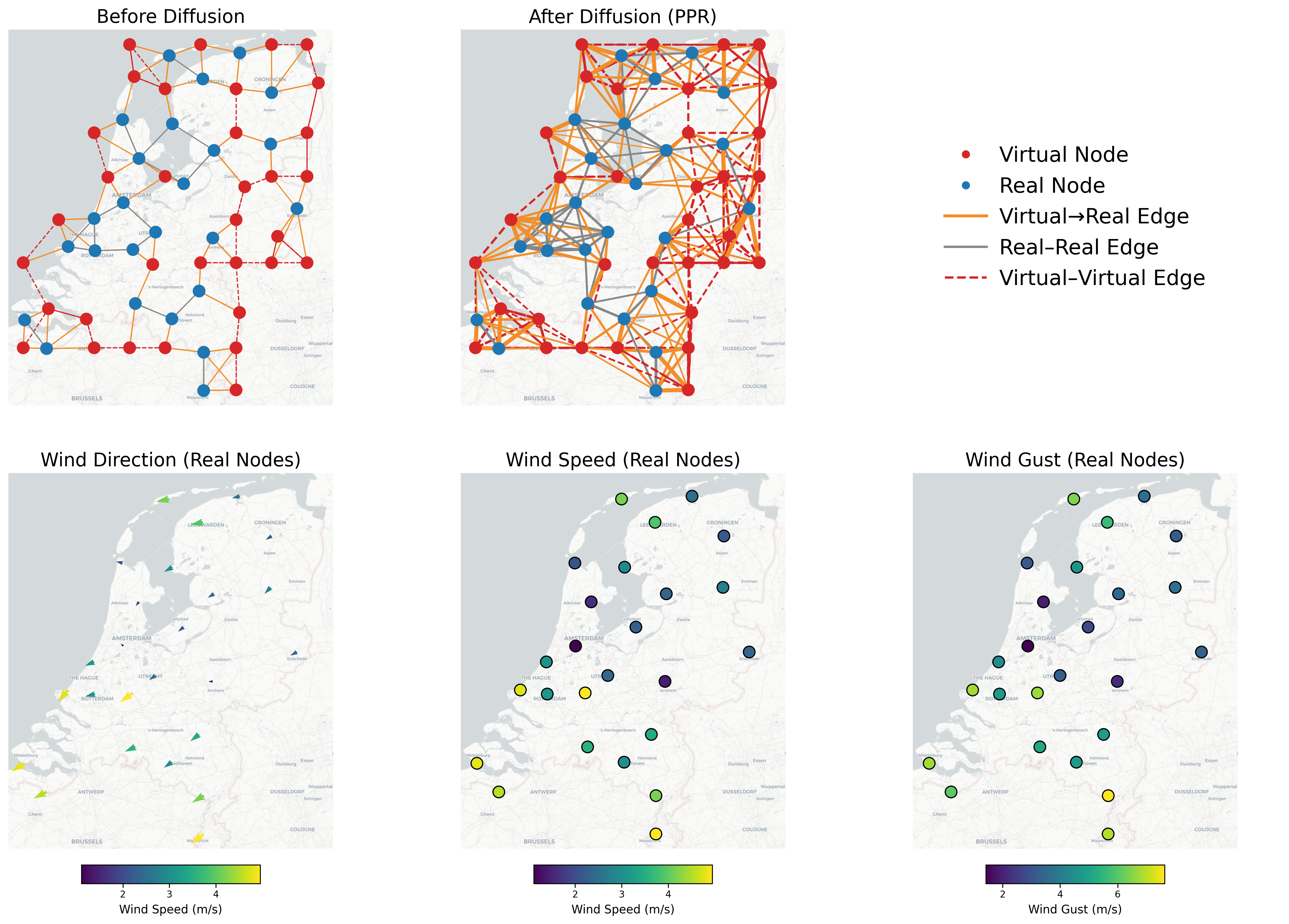}
        \label{fig:diffused_grapha}
    }\\[2mm]
    \subfloat[]{%
        \includegraphics[width=\textwidth]{influence_distribution.png}
        \label{fig:influence_distributionb}
    }
    \caption{Graph diffusion and virtual node dependence.
    (a) Diffusion process transforming the spatial graph structure and variable distribution across the Netherlands. 
    (b) Influence distribution showing how virtual nodes aggregate information from multiple nearby real stations.}
    \label{fig:graph_overview}
\end{figure}

\subsection{Evaluation}
To rigorously assess the performance of our proposed model, we conducted an evaluation based on predictions at eight specific virtual locations geographically aligned with existing eight meteorological stations. Model predictions at these eight test stations span the full temporal range covered by our dataset, enabling comprehensive temporal and spatial evaluation.

We compared model-generated nowcasts directly with the ground-truth observations obtained from the actual weather stations at these locations. To quantify the accuracy and reliability of the predictions, we utilized two widely recognized metrics: Mean Absolute Error (MAE) and Root Mean Square Error (RMSE) \cite{karunasingha2022root}. The MAE provides insight into the average absolute deviations of predictions from actual observations, offering a straightforward measure of forecast accuracy. RMSE, on the other hand, places greater emphasis on larger errors by squaring the differences, thus highlighting significant nowcasting inaccuracies. For wind direction, due to its circular nature, we employed angular versions of these metrics, namely Angular Mean Absolute Error (Angular MAE) and Angular Root Mean Square Error (Angular RMSE). These angular metrics capture the circular continuity and periodicity of directional data, thus providing a more accurate assessment for directional predictions.

These metrics were computed independently for each of the three critical wind-related variables considered in this study, wind direction, wind speed, and wind gust, to ensure a nuanced evaluation of the model’s predictive capabilities across different meteorological conditions. By evaluating the model's performance separately for each variable and aggregating results over the complete temporal span, we provided a detailed and robust assessment of our model’s generalization and practical utility in real-world nowcasting scenarios.

\section{Results}


\subsection{Information flow into unobserved regions}
To assess how the graph integrates real and virtual stations, we visualized its structure before and after diffusion (Fig.~\ref{fig:diffused_grapha}). Prior to diffusion, each virtual node was sparsely connected to its nearest neighbors. After applying personalized PageRank diffusion, the graph became denser, with stronger links emerging from virtual nodes to nearby real stations. This design passes information from observed to unobserved regions, while still permitting interactions among virtual nodes. Real stations retained their direct connections, preserving the coherence of the observed network. Spatial maps of wind direction, speed, and gust further illustrate how local meteorological variability shapes these diffusion patterns across the Netherlands.

We quantified the extent to which virtual nodes depend on neighboring real stations (Fig.~\ref{fig:influence_distributionb}). On average, only about one third of the incoming signal to a virtual node originated directly from real stations, with the remainder arriving indirectly through other virtual nodes. No single station dominated this influence, even the strongest neighbor typically contributed less than 20\%. Instead, the cumulative effect of the top three to five real neighbors accounted for most of the real signal, indicating that virtual nodes integrate information from multiple sources rather than relying on a single dominant station. Most virtual nodes received non-zero diffusion weights from two to three real stations, indicating that they typically integrate information from multiple neighbors rather than a single source. These results confirm that the diffusion design enables virtual nodes to blend signals from multiple observed stations, enhancing their representational stability and ensuring that spatial heterogeneity in the meteorological field is appropriately captured.

\subsection{Nowcasting accuracy}
\begin{figure*}
\centering

\begin{minipage}[t]{0.4\textwidth}\vspace{0pt}
  \centering
  \subfloat[]{%
    \includegraphics[width=\linewidth]{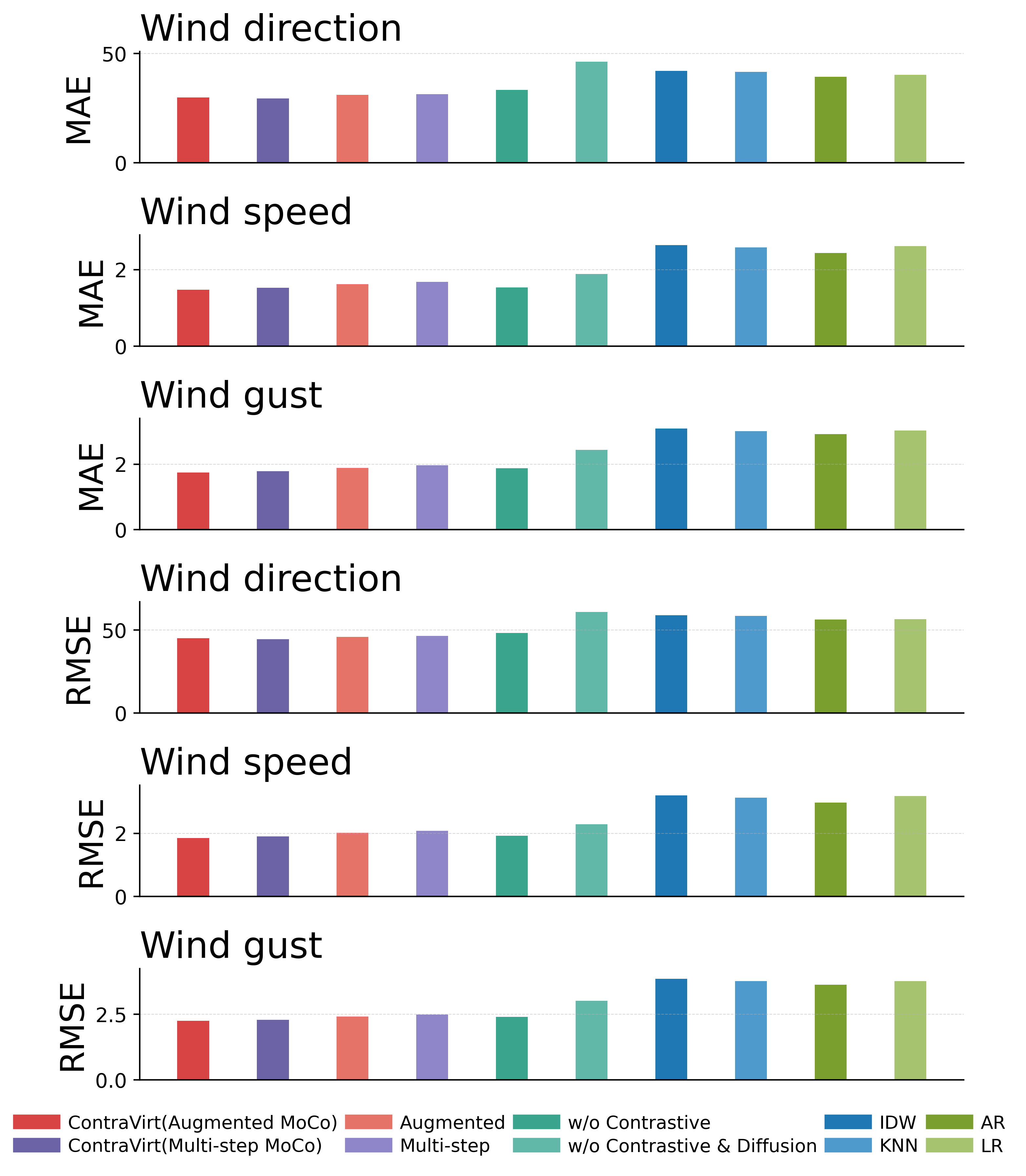}%
    \label{fig:leadtime:a}%
  }\\[0mm]
  \subfloat[]{%
    \includegraphics[width=\linewidth]{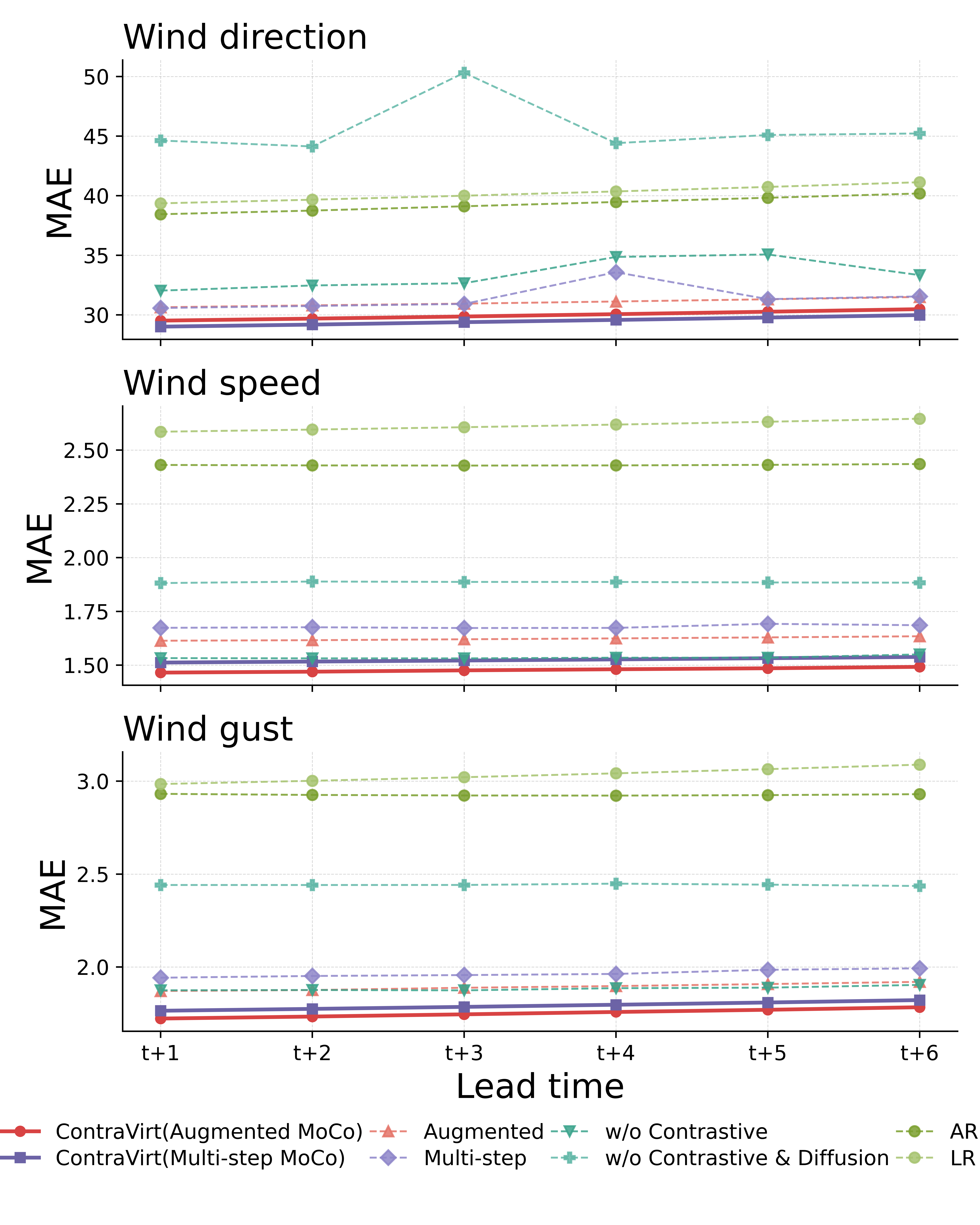}%
    \label{fig:leadtime:b}%
  }
\end{minipage}%
\hfill%
\begin{minipage}[t]{0.55\textwidth}\vspace{0pt}
  \centering
  \subfloat[]{%
    \includegraphics[width=\linewidth]{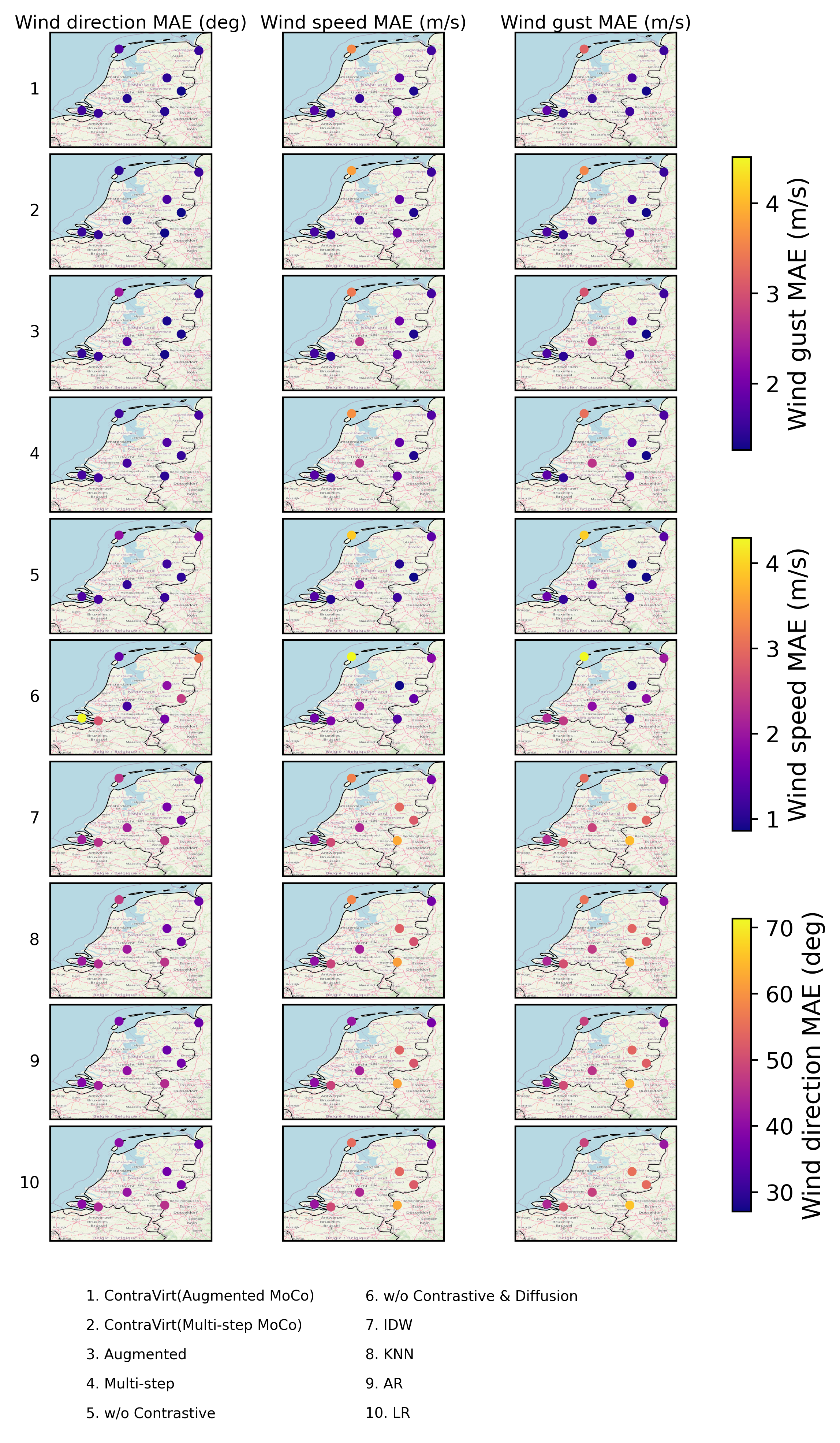}%
    \label{fig:leadtime:c}%
  }
\end{minipage}

\caption{Model performance across metrics, lead times, and test stations.
(a) MAE/RMSE across models and targets. (b) MAE in different lead time ($t{+}1$–$t{+}6$).
(c) Spatial distribution of MAE across the eight testing stations (10 variants). Our methods achieve lower short-lead errors and degrade more slowly, especially Augmented MoCo.}
\label{fig:leadtime}
\end{figure*}

We compared our models with both ablated variants and traditional baselines across three wind variables (Fig. \ref{fig:leadtime:a}; Table \ref{tab:3}). Models combining contrastive learning with diffusion consistently outperformed all baselines. ContraVirt(Multi-step MoCo) achieved the lowest error for wind direction (MAE = 29.49°, RMSE = 44.52°), while ContraVirt(Augmented MoCo) performed best for wind speed (MAE = 1.48 m/s, RMSE = 1.85 m/s) and gusts (MAE = 1.75 m/s, RMSE = 2.25 m/s). These results represent error reductions of  26–29\% over regression baselines (AR, LR) for wind direction, and over 40\% for wind speed and gust. Relative to interpolation baselines (IDW, KNN), the improvements reach 29\% for direction and 43\% for speed and gust.

\begin{table}
\centering
\renewcommand{\arraystretch}{1.15}
\setlength{\tabcolsep}{3pt}
\caption{Forecasting performance (MAE and RMSE) across models.Models are evaluated on wind direction (°), wind speed, and wind gust (m/s).}
\label{tab:3}
\begin{tabular}{lcccccc}
\toprule
\multirow{2}{*}{\textbf{Model}} & \multicolumn{3}{c}{\textbf{MAE}} & \multicolumn{3}{c}{\textbf{RMSE}} \\
\cmidrule(lr){2-4} \cmidrule(lr){5-7}
& \textbf{Direction} & \textbf{Speed} & \textbf{Gust} & \textbf{Direction} & \textbf{Speed} & \textbf{Gust} \\
\midrule
ContraVirt (Augmented MoCo) & 29.98 & \textbf{1.48} & \textbf{1.75} & 45.16 & \textbf{1.85} & \textbf{2.25} \\
ContraVirt (Multi-step MoCo) & \textbf{29.49} & 1.52 & 1.79 & \textbf{44.52} & 1.91 & 2.29 \\
Augmented & 31.06 & 1.62 & 1.89 & 45.94 & 2.02 & 2.41 \\
Multi-step & 31.45 & 1.68 & 1.97 & 46.45 & 2.08 & 2.50 \\
\midrule
w/o Contrastive & 33.41 & 1.54 & 1.88 & 48.37 & 1.92 & 2.41 \\
w/o Contrastive \& Diffusion & 46.30 & 1.89 & 2.44 & 61.00 & 2.29 & 3.02 \\
\midrule
AR & 39.30 & 2.43 & 2.93 & 56.40 & 2.98 & 3.63 \\
LR & 40.21 & 2.61 & 3.03 & 56.69 & 3.18 & 3.76 \\
KNN & 41.57 & 2.58 & 3.01 & 58.57 & 3.13 & 3.76 \\
IDW & 42.00 & 2.64 & 3.10 & 58.96 & 3.21 & 3.86 \\
\bottomrule
\end{tabular}
\end{table}

Ablation experiments highlighted the critical role of contrastive learning and graph diffusion. Eliminating the contrastive objectives increased errors across all variables (e.g., wind direction MAE rising from 29.5° to 33.4°), while removing both contrastive learning and diffusion caused a marked collapse in performance (wind direction MAE = 46.3°, RMSE = 61.0°). These results demonstrate that contrastive alignment and graph diffusion are essential for robust representation learning in data-scarce environments.

Classical methods including auto-regression (AR), linear regression (LR), KNN, and IDW produced substantially larger errors. For instance, IDW yielded a wind speed MAE of 2.64 m/s and a gust MAE of 3.10 m/s, nearly double those of our ContraVirt models. These results highlight the inability of interpolation and simple regression to capture spatial heterogeneity, and demonstrate that our contrastive diffusion framework provides more accurate and stable forecasts across all wind variables.

We further evaluated performance across forecast horizons (Fig. \ref{fig:leadtime:b}). Although all models exhibited increasing errors with lead time, ContraVirt consistently maintained lower error rates, particularly at longer horizons. ContraVirt(Augmented MoCo) achieved the most accurate short-term predictions ($t+1$–$t+2$) and degraded more gracefully over six steps. In contrast, non-contrastive and baseline methods accumulated errors more rapidly, especially for wind direction, where their divergence from our model was most pronounced at extended horizons.

We also evaluated all models at the eight independent test stations distributed across the Netherlands (Fig. \ref{fig:leadtime:c}). This station-level analysis offers a spatial perspective on generalization beyond the real stations. ContraVirt(Augmented MoCo) and ContraVirt(Multi-step MoCo) consistently achieved the lowest errors across nearly all sites and variables. The largest improvements were observed at coastal and northern locations, where observational coverage is sparse and interpolation baselines performed poorly. These results show that contrastive diffusion not only reduces overall error but also delivers geographically consistent gains, maintaining forecast reliability even in the most data-scarce regions.

\subsection{Nowcasting across seasons}

\begin{figure*}
  \centering

  \subfloat[]{%
    \includegraphics[width=0.95\textwidth]{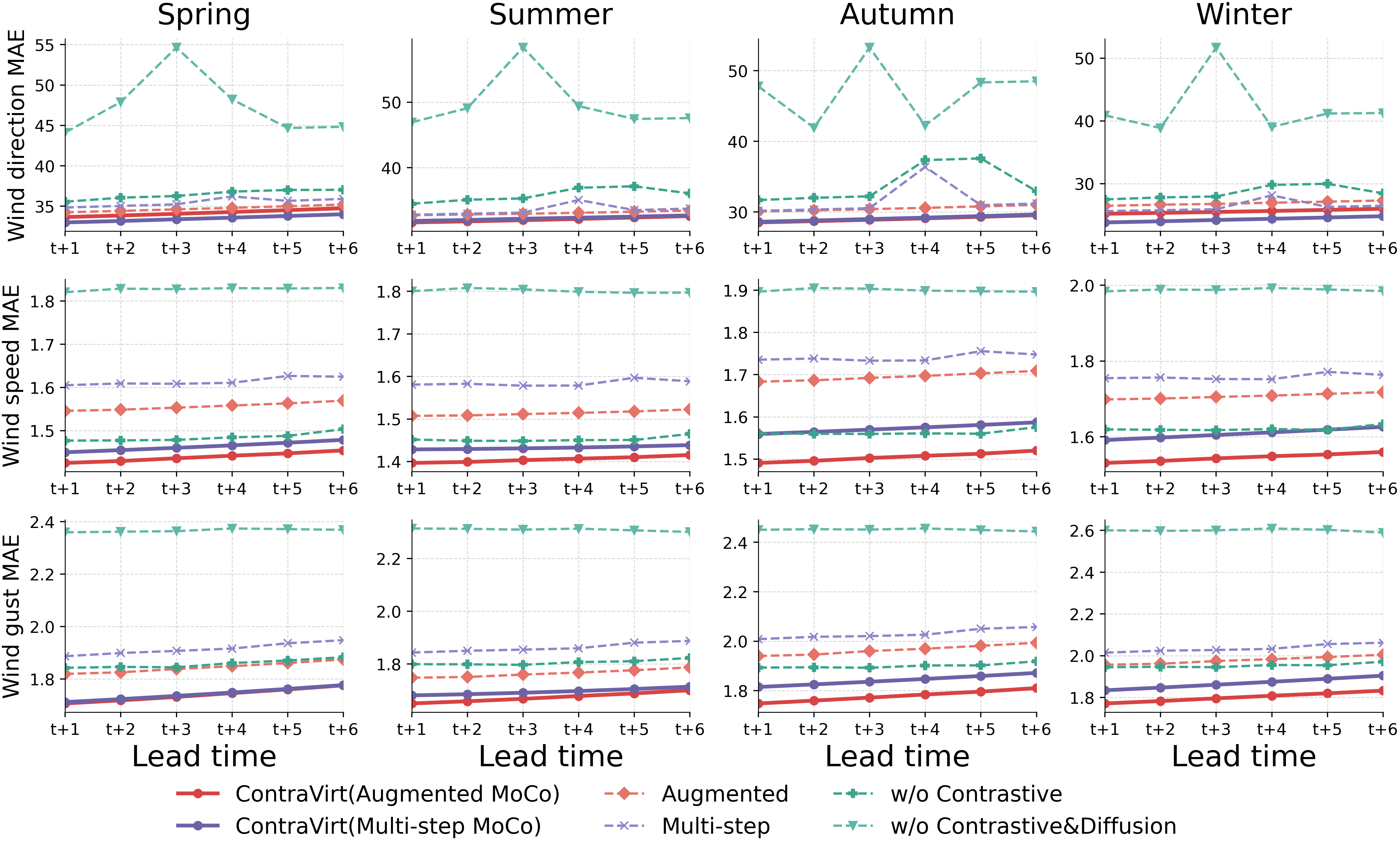}%
    \label{fig:season-a}%
  }\\[0mm]

  \subfloat[]{%
    \includegraphics[width=0.95\textwidth]{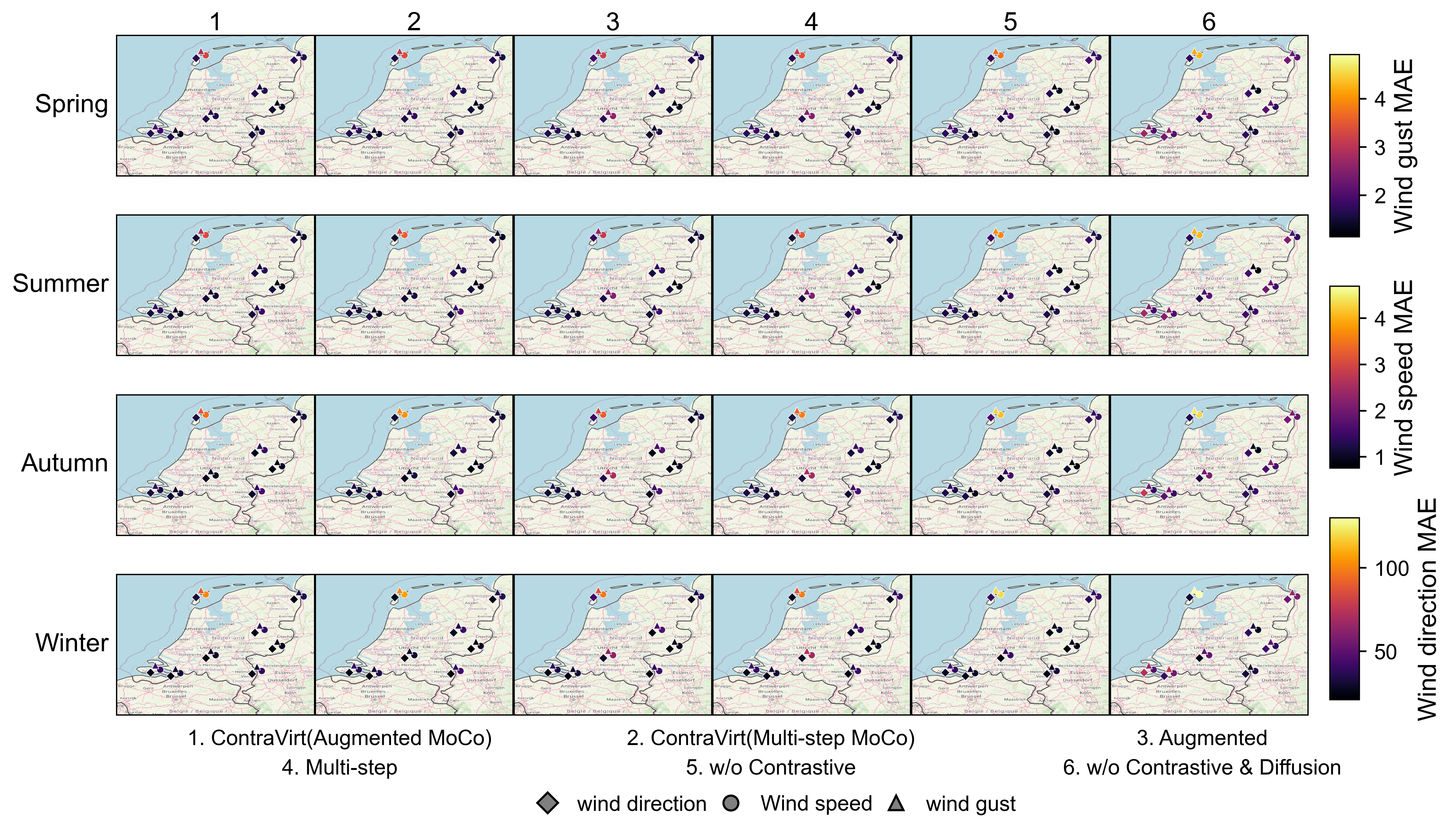}%
    \label{fig:season-b}%
  }

  \caption{Seasonal MAE results.
  (a) Lead–time MAE for wind direction, wind speed, and wind gust.
  (b) Station-wise MAE patterns and seasonal variations across the Netherlands.}
  \label{fig:season-four}
\end{figure*}

\begin{figure*}
  \centering
  \subfloat[]{%
    \includegraphics[width=0.95\textwidth]{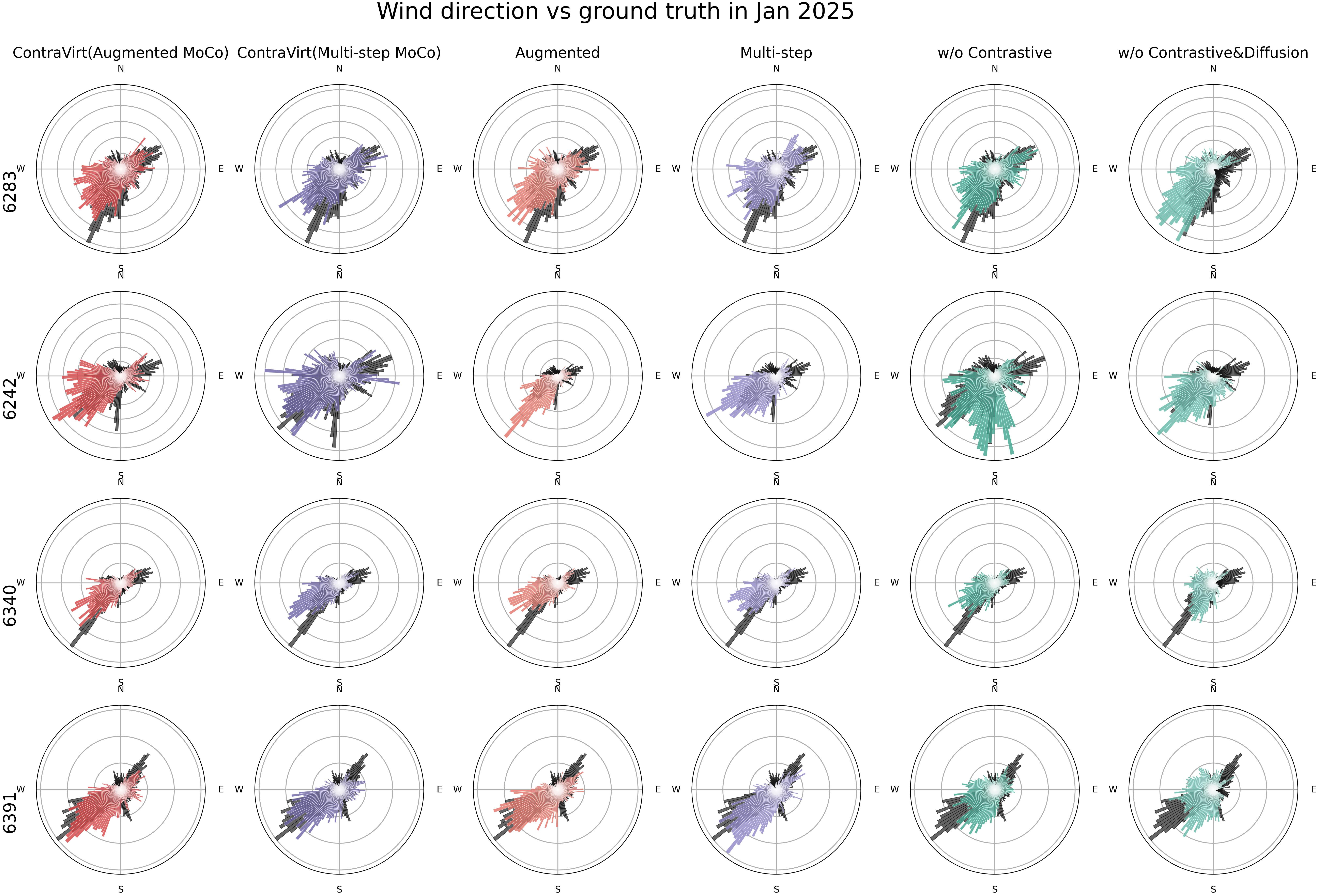}%
    \label{fig:rose_a}%
  }\\[3mm]
  
  \subfloat[]{%
    \includegraphics[width=0.47\textwidth]{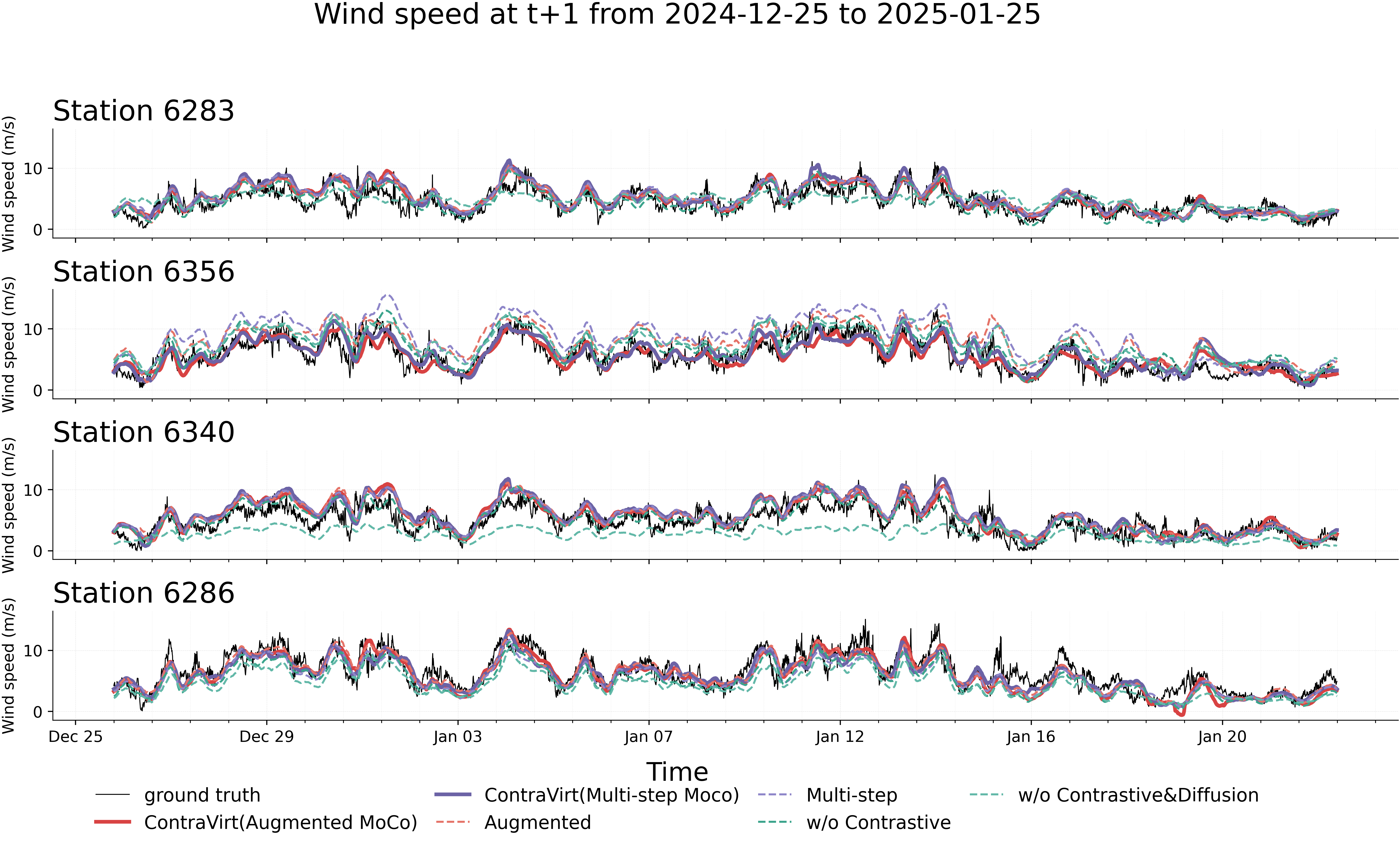}%
    \label{fig:rose_b}%
  }\hfill
  \subfloat[]{%
    \includegraphics[width=0.47\textwidth]{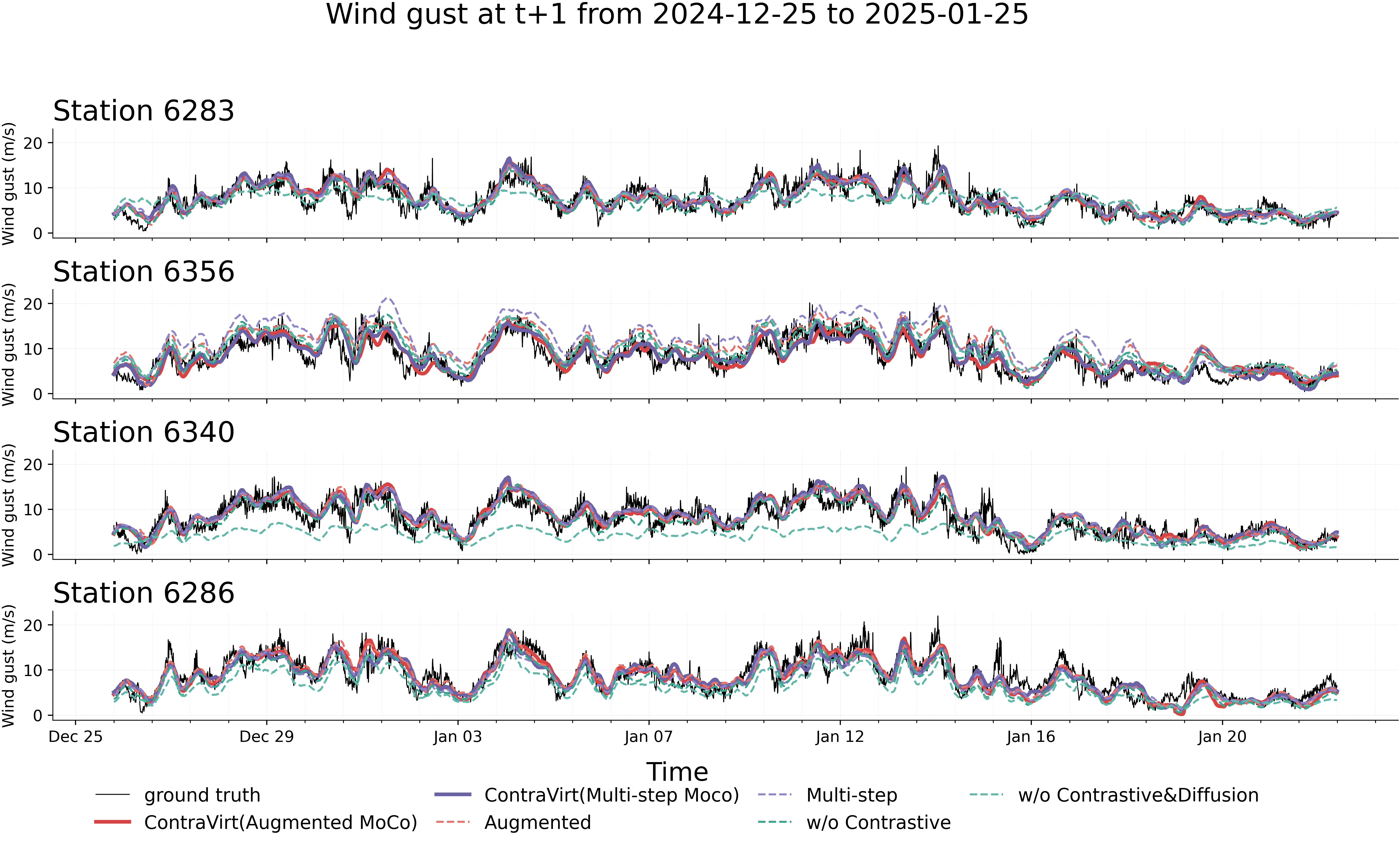}%
    \label{fig:rose_c}%
  }

  \caption{Case studies of short-horizon wind forecasts.
  (a) Wind-direction roses at horizon $t{+}1$ for four representative test stations (6283, 6242, 6340, 6391), comparing ground truth (black) with six models during a storm-prone winter month in the Netherlands. 
  (b,c) Time–series comparisons of wind speed and wind gust at the same stations, showing close alignment between predictions and observations.}
  \label{fig:case_study}
\end{figure*}

Seasonal analysis further highlighted the strengths of our approach under the Netherlands’ highly variable wind climate (Fig. \ref{fig:season-a}). Wind direction strongly influenced by rapidly shifting air masses and heterogeneous surface effects, remained the most difficult variable to predict \cite{gryning2020investigating}. However, ContraVirt model consistently outperformed ablation variants and traditional baselines across all seasons and lead times. Wind direction errors remained close to 30°, compared with values exceeding 50° when contrastive learning or diffusion was removed. This finding is consistent with prior work showing that wind direction is more challenging to forecast than wind speed in the Netherlands \cite{gryning2020investigating}. Across all four seasons, ContraVirt(Multi-step MoCo) delivered the lowest errors, with the largest improvements during autumn and winter, when storms and extreme gusts are most frequent. In contrast, removing contrastive learning or diffusion caused marked performance declines, most notably for wind direction, where errors often rose above 45–50°. These results demonstrate that both components are essential for capturing rapid shifts driven by pressure changes and coastal effects.

Wind speed forecasts showed smaller seasonal variability than gusts or direction, reflecting their stronger temporal persistence (Fig. \ref{fig:season-a}). Nonetheless, our models retained clear advantages across lead times, with error growth remaining limited even during winter’s energetic mid-latitude cyclones (Fig. \ref{fig:season-b}). Wind gusts proved the most difficult to predict; however, contrastive diffusion substantially reduced errors compared to ablations and baselines, especially during storm seasons when turbulent bursts are most frequent \cite{baatsen2015severe}.

Seasonal evaluations further highlight the difficulty of nowcasting wind gusts in the Netherlands. As shown in the lead-time errors (Fig. \ref{fig:season-a}) and station-wise MAE maps (Fig. \ref{fig:season-b}), gust predictions consistently exhibit higher errors than wind speed, with the largest values in autumn and winter. These peaks coincide with the storm season, when extratropical cyclones dominate the wind climate and drive the most extreme gusts. Previous studies have shown that, unlike convective gusts in summer, the strongest gusts in the Netherlands are typically generated by large-scale extratropical storms in winter \cite{wever2009improving}. Projections further suggest that rising sea surface temperatures may allow tropical cyclones to track farther north, reaching the Netherlands as ex-tropical systems carrying damaging winds \cite{baatsen2015severe}. Together, these results underscore the physical consistency of our findings, showing that our models capture systematic seasonal challenges in gust prediction that align with known atmospheric drivers.

Spatial MAE maps (Fig. \ref{fig:season-b}) also demonstrated that improvements were not confined to specific locations. Although coastal and northern stations remained the most difficult to forecast, our model generalized well across both inland and maritime regions. This spatial robustness highlights the model’s capacity to exploit relational structure across stations, extending predictive skill into unobserved areas where local observations are sparse.

\begin{figure}
\centering

\subfloat[]{%
  \includegraphics[width=0.5\linewidth]{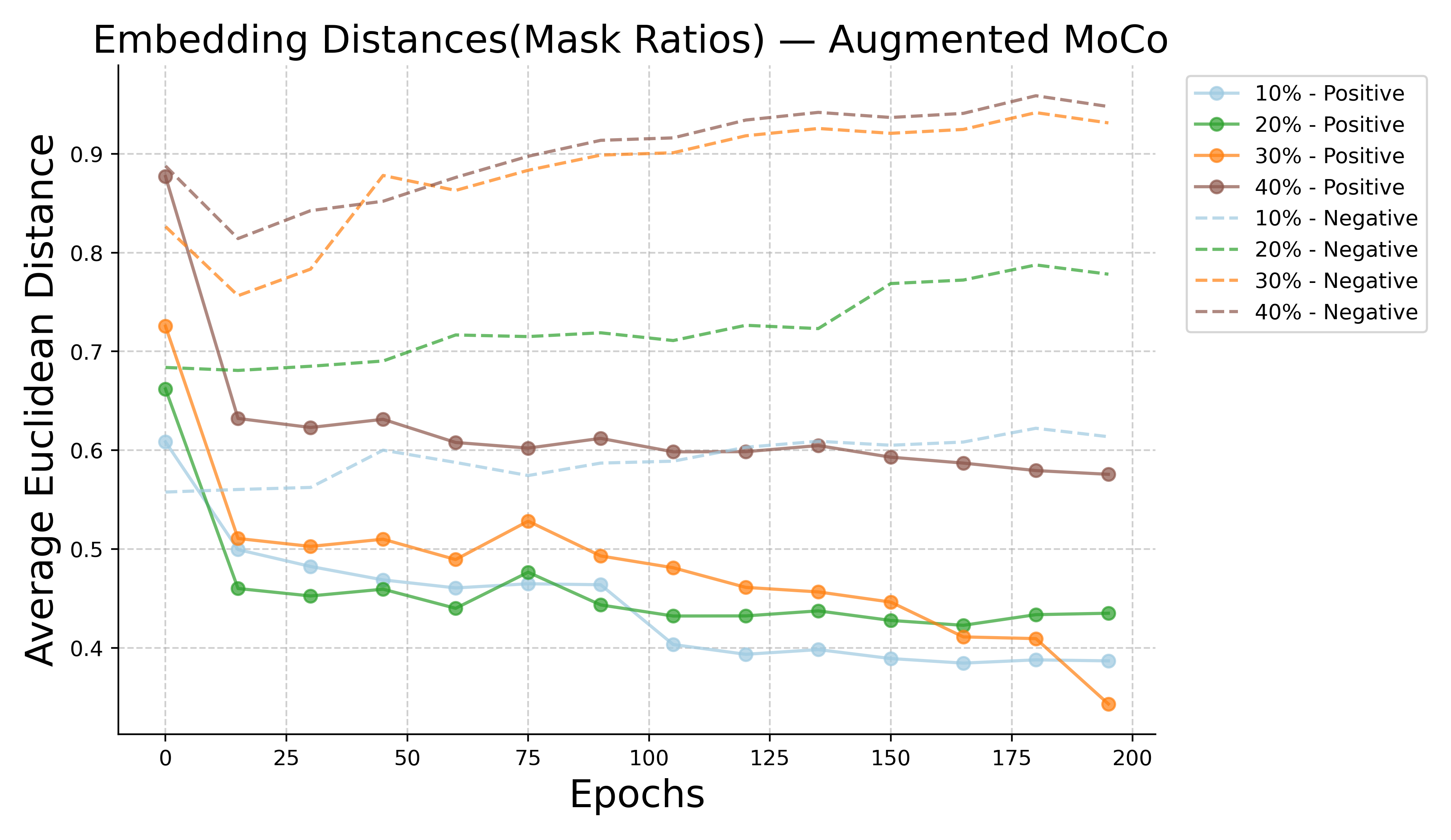}%
  \label{fig:augmented_moco_curve}%
}
\hfill
\subfloat[]{%
  \includegraphics[width=0.5\linewidth]{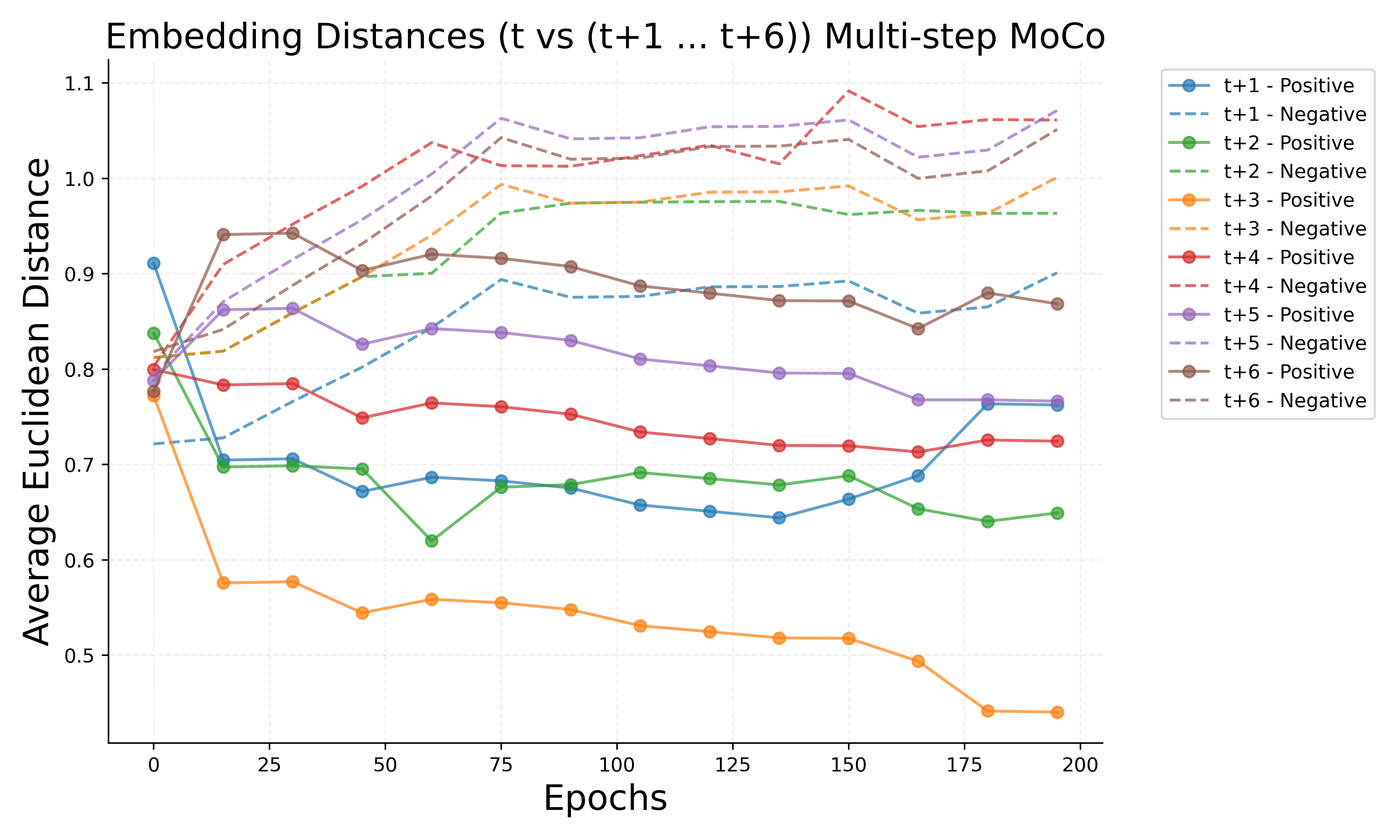}%
  \label{fig:multi_step_moco_curve}%
}

\vspace{3mm}

\subfloat[]{%
  \includegraphics[width=0.5\linewidth]{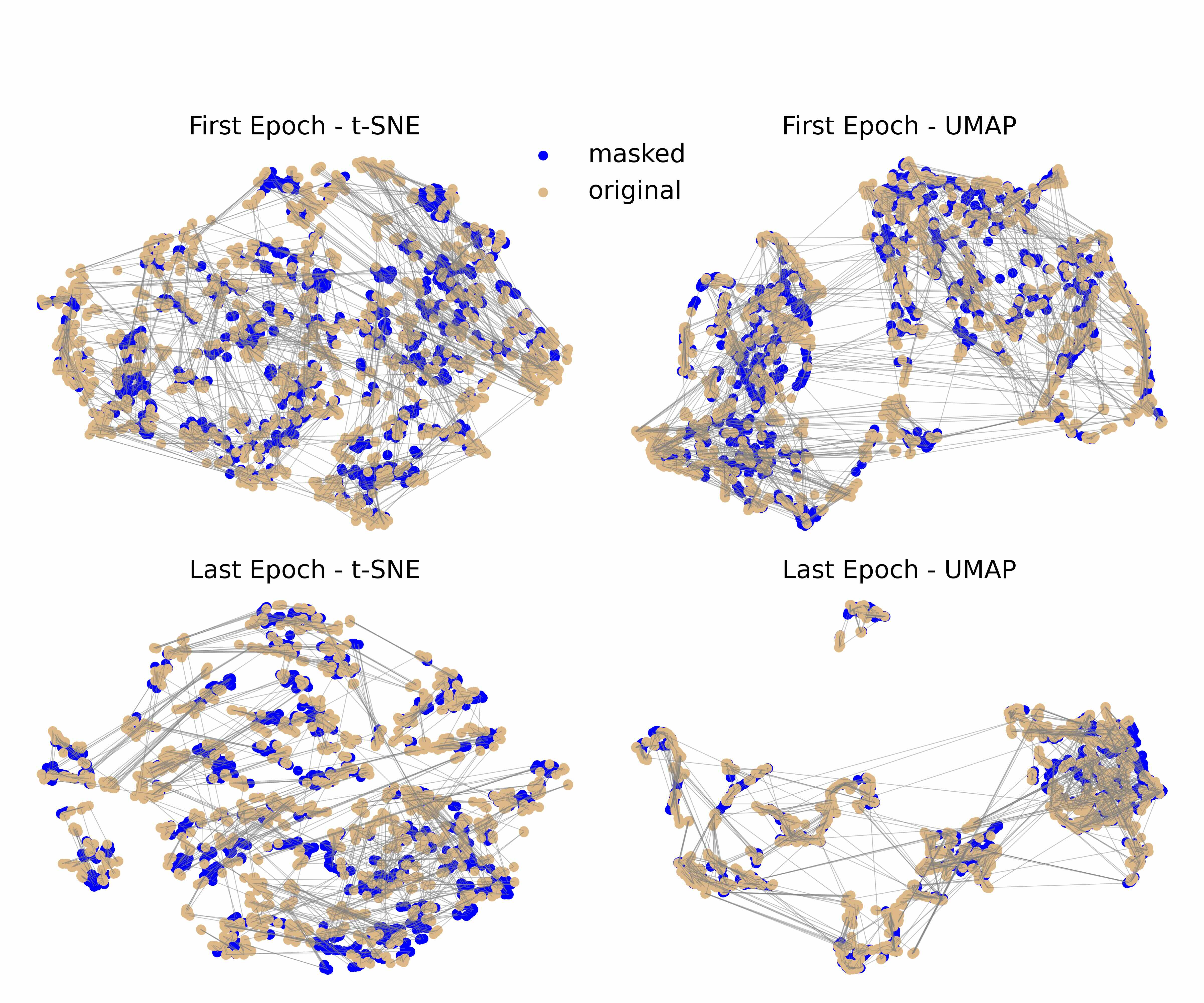}%
  \label{fig:augmented_moco_proj}%
}
\hfill
\subfloat[]{%
  \includegraphics[width=0.5\linewidth]{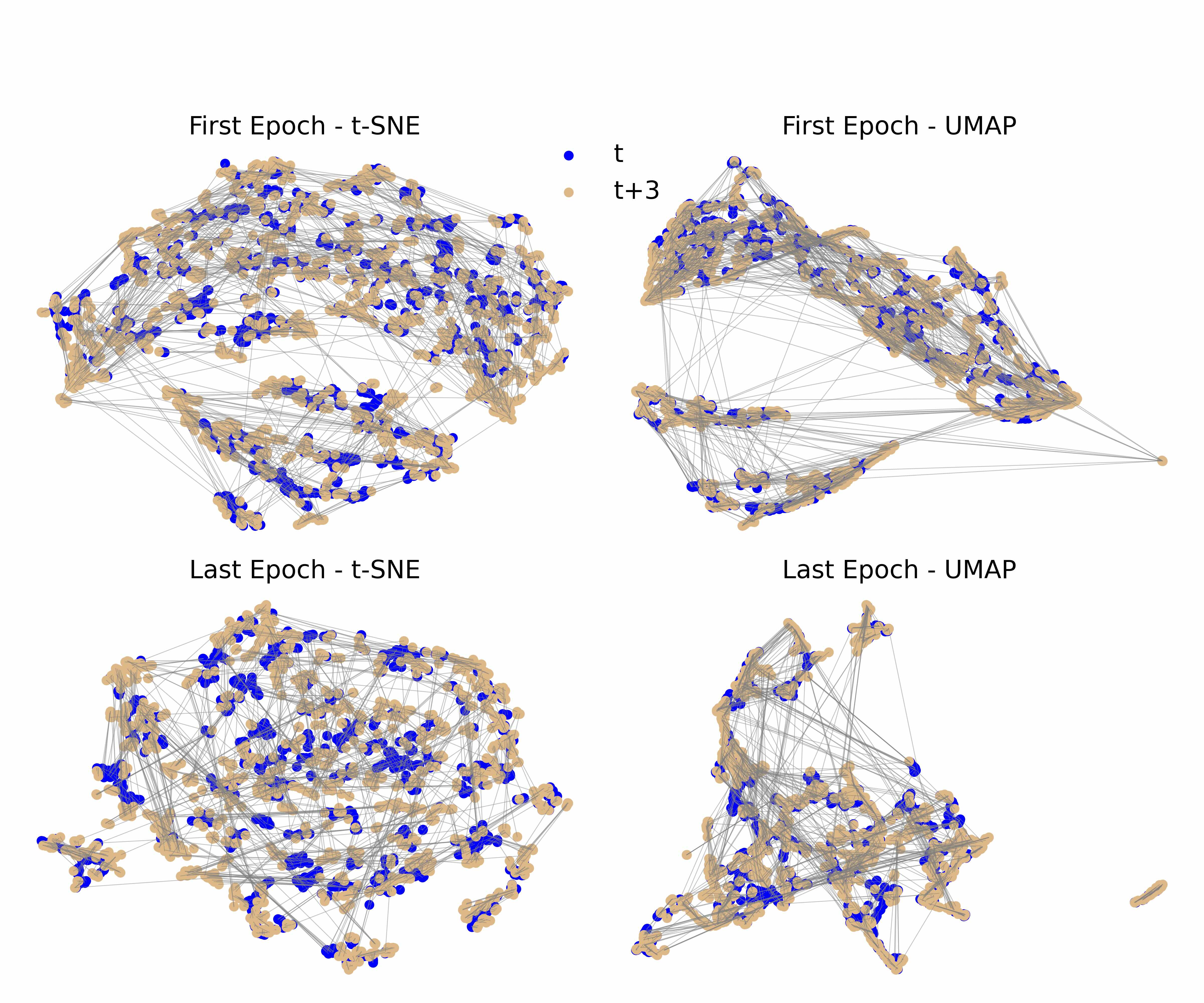}%
  \label{fig:multi_step_proj}%
}

\caption{Embedding distance dynamics and projections.
(a,b) Positive and negative pair-distance evolution for augmented and multi-step MoCo. 
(c,d) t-SNE and UMAP projections showing progressively tighter contrastive embeddings.}
\label{fig:embedding_curves_and_projections}
\end{figure}

\subsection{Station-level evaluation}

We next evaluated station-level performance for wind direction (Fig. \ref{fig:rose_a}). Wind-direction roses provide a distributional comparison of predicted and observed directions. The ContraVirt(Multi-step MoCo) model reproduced both the observed directional spread and the predominant south-westerlies with greater fidelity, whereas baseline models underestimated variability and produced overly narrow predictions. This distributional accuracy is particularly important in the Netherlands, where prevailing westerly flows, often south-westerly, arise from the interaction of maritime and continental air masses, generating highly dynamic wind regimes \cite{cheneka2023quantifying}.

Time series of wind speed (Fig. \ref{fig:rose_b}) and gust (Fig. \ref{fig:rose_c}) further illustrate model skill in capturing daily variability. ContraVirt(Augmented MoCo) closely tracked the timing and magnitude of gust peaks, alleviating the underestimation of extremes evident in non-contrastive and interpolation baselines. Together, these station-level analyses highlight the dual strength of our framework: capturing the full directional distribution of winds while preserving the temporal fidelity of speed and gust forecasts.

\subsection{Contrastive embedding dynamics}

To evaluate the choice of contrastive strategies, we first varied the masking ratio in the Augmented MoCo setting from 10\% to 40\% (Fig. \ref{fig:augmented_moco_curve}). Light masking (10–20\%) preserved too much information, yielding only modest alignment between masked and original embeddings. By contrast, overly aggressive masking (40\%) reduced the margin between positive and negative pairs. A ratio of 30\% achieved the clearest separation, indicating that moderate information removal is essential for learning robust invariances without overwhelming the model’s reconstruction capacity.

We next analyzed embedding distances across epochs in the Multi-step MoCo framework (Fig. \ref{fig:multi_step_moco_curve}). Positive pairs were defined as a virtual node and its geographically nearest real neighbor at the future time step, all remaining combinations served as negative samples. Among the six temporal offsets, the $t$ vs. $t+3$ contrast produced the strongest margin. The positive pairs stabilized at smaller distances, while negatives increased, maximizing discrimination. Very short offsets (e.g., $t+1$, 10 minutes) showed weak separation, reflecting minimal atmospheric change within such brief intervals, whereas long offsets (e.g., $t+6$, 1 hour) weakened temporal correlation. These results suggest that an intermediate offset of three steps (30 minutes) provides the optimal balance between persistence and variability, yielding the most effective supervisory signal for representation learning.

Visualization of the learned embedding space using t-SNE \cite{cai2022theoretical} and UMAP \cite{ghojogh2021uniform} further confirmed these dynamics (Fig. \ref{fig:augmented_moco_proj} and Fig.\ref{fig:multi_step_proj}). At initialization, masked and unmasked embeddings were poorly aligned, showing wide dispersion and little overlap. By the end of training, positive pairs converged to near positions, and their neighborhoods became tightly organized, while negatives remained dispersed. The convergence was especially clear for the 30\% masking strategy, where masked–original embeddings collapsed into coherent clusters. For Multi-step MoCo, early-stage embeddings exhibit substantial spatial dispersion, indicating limited temporal consistency. After contrastive training, the embeddings become more aligned and compact, reflecting effective learning of temporal features and greater representational stability.

These analyses demonstrate that contrastive alignment benefits most from intermediate temporal offsets ($t$ vs $t+3$) and moderate masking (30\%). These design choices maximize temporal discrimination while enforcing spatial invariance, providing strong self-supervised representation learning in nodes without direct observations.

\section{Conclusion}
Accurate wind nowcasting in unobserved regions remains a long-standing challenge with direct implications for energy, safety, and environmental planning. ContraVirt addresses this gap by integrating real meteorological stations with virtual nodes, enabling nowcasts where no direct observations exist. By combining diffusion-based message passing with contrastive learning, the model consistently outperforms interpolation and regression baselines, delivering both higher accuracy and stronger generalization in unobserved regions. 
The implementation of our models is available at 
\href{https://github.com/JieJieNiu/-Wind-Nowcasting-in-Unobserved-Regions}{GitHub}.


\bibliographystyle{elsarticle-num}
\bibliography{Main}

\end{document}